\documentclass[A4paper,12pt]{article}

\usepackage{bm}
\usepackage{graphicx}
\usepackage{amssymb}  
\usepackage{amsthm}
\usepackage{amsmath}

\usepackage[T1]{fontenc}    
\usepackage{hyperref}       
\usepackage{url}            
\usepackage{booktabs}       
\usepackage{amsfonts}       
\usepackage{nicefrac}       
\usepackage{microtype}      

\usepackage{algorithmicx}
\usepackage[ruled] {algorithm2e}
\usepackage{subfigure}
\usepackage{caption}

\newtheorem{theorem}{\sc Theorem}[section]
\newtheorem{lemma}[theorem]{\sc Lemma}

\newtheorem{proposition}[theorem]{\sc Proposition}

\newtheorem{definition}[theorem]{\sc Definition}

\begin{document}

\title{Optimal Learning for Sequential Decision Making for Expensive Cost
Functions with Stochastic Binary Feedbacks }

\author{
{\sc Yingfei Wang}
\thanks{Department of Computer Science, Princeton University, Princeton, NJ 08540, yingfei@cs.princeton.edu}
\and
  {\sc Chu Wang}
\thanks{The Program in Applied and Computational Mathematics, Princeton University, Princeton, NJ 08544 }
\and
{\sc Warren Powell}
\thanks{Department of Operations Research and Financial Engineering, Princeton University, Princeton, NJ 08544}
}
\date{}
\maketitle

\begin{abstract}
We consider the problem of sequentially making decisions that are rewarded by ``successes" and ``failures" which can be predicted through an unknown relationship that depends  on a partially
controllable vector of attributes for each instance. The learner takes an active role in selecting samples from the instance pool. The goal is to maximize the probability of success in either offline (training) or online (testing) phases. Our problem is motivated by real-world applications where observations are time consuming and/or expensive. We develop a knowledge gradient policy using an online Bayesian linear classifier to guide the experiment by maximizing the expected value of information of labeling each alternative. We provide a finite-time analysis of the estimated error and show that the maximum likelihood estimator based produced by the KG policy is consistent and asymptotically normal. We also show that the knowledge gradient policy is asymptotically optimal in an offline setting.  This work further extends the knowledge gradient to the setting of contextual bandits. We report the results of a
series of experiments that demonstrate its efficiency.
\end{abstract}


%


\section{Introduction}
There are many real-world optimization tasks where observations
are time consuming and/or expensive. This often occurs in experimental sciences where testing different values of controllable parameters of a physical system is expensive.  Another example
arises in  health services,  where physicians have to
make medical decisions (e.g. a course of drugs, surgery,
and expensive tests) and  we can characterize an outcome as a success (patient does not need to return
for more treatment) or a failure (patient does need
followup care such as repeated operations).   This puts us in the setting of optimal learning where the number of objective function samples are limited, requiring that we
learn from our decisions as quickly as possible. This represents a distinctly different learning environment than what has traditionally been considered using popular policies such as upper confidence bounding (UCB) which have proven effective in fast-paced settings such as learning ad-clicks.

In this paper, we are interested in a sequential decision-making setting where at each time step we choose one of finitely many decisions and observe a stochastic binary reward where each instance is described by various controllable and uncontrollable attributes. The goal is to maximize the probability of success in either offline (training) or online (testing) phases. There are a number of applications that can easily fit into our success/failure model:
\begin{itemize}
\item Producing single-walled nanotubes.  Scientists have physical procedures to produce nanotubes. It can produce either single-walled or double walled nanotubes through an unknown relationship with the  controllable parameters, e.g. laser poser, ethylene, Hydrogen and pressure. Yet only the single-walled nanotubes are acceptable. The problem is  to quickly learn the best parameter values with the highest probability of success (producing single-walled nanotubes).
\item Personalized health care.  We consider the problem of how to choose clinical pathways (including surgery, medication and tests) for different upcoming patients to maximize the success of the treatment.  
\item Minimizing the default rate for loan applications. When facing borrowers with different background information and credit history, a lending company needs to decide whether to grant a loan, and  with what terms (interest rate, payment schedule).
\item Enhancing the acceptance of the admitted students.  A university needs to decide which students to admit and how much aid to offer so that the students  will then accept the offer of admission and matriculate.  
\end{itemize}

Our work will initially focus on offline settings such as laboratory experiments or medical trials where we are not punished for errors incurred during training and instead are only  concerned with the final recommendation after the offline training phases. We then extend our discussion to online learning settings where the goal is to minimize cumulative regret. We also consider  problems with partially controllable attributes, which is known as contextual bandits. For example, in the health care problems, we do not have control over  the patients (which is represented by  a feature vector including demographic characteristics, diagnoses, medical history) and can only  choose the medical decision. A university cannot control which students are applying to the university.   When deciding whether to grant a loan, the lending company cannot choose the personal information of the borrowers.

Scientists can draw on an extensive body of literature on the classic design of experiments \cite{morris1970optimal, wetherill1986sequential, montgomery2008design} where the goal is to decide which observations to make when fitting a function. Yet in our setting, the decisions are guided by a well-defined utility function (that is,  maximize  the probability of success).    The problem is related to the literature on active learning \cite{schein2007active, tong2002support, freund1997selective, settles2010active}, where our setting is most similar to membership query synthesis where the learner may request labels
for any unlabeled instance in the input space to  learn 
a classifier that accurately predicts the labels of new examples. By contrast, our goal is to maximize a utility function such as the success of a treatment.  Moreover, the expense of labeling each alternative sharpens the conflicts of learning the prediction and finding the best alternative.  

Another similar sequential decision making setting is multi-armed bandit problems  \cite{auer2002finite,bubeck2012regret, filippi2010parametric,mahajan2012logucb, srinivas2009gaussian,chapelle2011empirical,wang2017efficient}.  Different belief models have been studied under the name of contextual bandits, including linear models \cite{chu2011contextual} and Gaussian process regression \cite{krause2011contextual}.    The focus of bandit work is minimizing cumulative regret in
an online setting, while we consider the performance of the final recommendation after an offline training phase.  There are recent works to address the problem we describe here by minimizing the simple regret. But first, the UCB type policies \cite{audibert2010best} are not best suited for expensive experiments. Second, the work on simple regret minimization  \cite{hoffman2014correlation, hennig2012entropy} mainly focuses on real-valued functions  and do not consider the problem with stochastic binary feedbacks.

There is  a literature on Bayesian optimization   \cite{he2007opportunity,chick2001new,powell2012optimal}. Efficient global optimization (EGO), and related methods such as sequential kriging optimization (SKO) \cite{jones1998efficient, huang2006global}  assume a Gaussian process belief model which does not scale to the higher dimensional settings that we consider. Others assume lookup table, or low-dimensional parametric methods, e.g. response surface/surrogate models \cite{gutmann2001radial, jones2001taxonomy,regis2005constrained}.  The existing literature mainly focuses on real-valued functions and none of these methods are directly suitable for our problem of maximizing the probability of success with binary outcomes.   A particularly relevant body of work in the Bayesian optimization literature is the expected improvement (EI) for binary outputs \cite{tesch2013expensive}. Yet when EI decides which alternative to measure, it is based on the expected improvement over current predictive posterior distribution while ignoring the potential change of the posterior distribution resulting from the next stochastic measurement (see Section 5.6 of \cite{powell2012optimal} and \cite{huang2006global} for detailed explanations).

We investigate a knowledge gradient policy that maximizes the value of information, since this approach is particularly well suited to problems where observations are expensive.  After its first appearance for ranking and selection problems \cite{frazier2008knowledge}, KG has been extended to various other belief models (e.g.  \cite{mes2011hierarchical,negoescu2011knowledge,wang2015nested}). Yet there is no KG variant designed for binary classification with  parametric belief models. In this paper, we extend the KG policy to the setting of classification problems under a Bayesian classification belief model which introduces the computational challenge of working with nonlinear belief models. 

This paper makes the following contributions. 1. Due to the sequential nature of the problem,  we  develop a fast online Bayesian linear classification procedure based on Laplace approximation for general link functions to recursively predict the response of each alternative. 2. We design a knowledge-gradient type policy for stochastic binary responses to guide the experiment. It can work with any choice of link function $\sigma(\cdot)$ and approximation procedure.  3. Since the knowledge gradient policy adaptively chooses the next sampled points, we provide a finite-time analysis on the estimated error that does not rely on the i.i.d. assumption. 4. We show that the maximum likelihood estimator based on the adaptively sampled points by the KG policy is consistent and asymptotically normal. We show furthermore that the knowledge gradient policy is asymptotic optimal. 5. We extend the KG policy to contextual bandit settings with stochastic binary outcomes. 

The paper is organized as follows. In Section \ref{sec:problem}, we establish the mathematical model for the problem of sequentially maximizing the response under binary outcomes.   We give a brief overview of (Bayesian) linear classification in Section \ref{BLR}.  We  develop an online Bayesian linear classification procedure based on Laplace approximation  in Section \ref{sec:oblc}. In Section \ref{sec:KG}, we propose a knowledge-gradient type policy for stochastic binary responses with or with context information. We theoretically study its finite-time and asymptotic behavior. Extensive demonstrations and comparisons of methods  for offline objective are demonstrated in Section \ref{EXP}.   

\section{Problem formulation} \label{sec:problem}

We assume that we have a finite set of alternatives $\bm{x}\in \mathcal{X}=\{\bm{x}_1,\dots,\bm{x}_M\}$. The observation of measuring each $\bm{x}$ is a binary outcome $y \in \{-1,+1\}$/\{failure, success\} with some unknown probability $ p(y=+1|\bm{x})$.   The
learner sequentially chooses a series of points  $(\bm{x}^0,\dots,\bm{x}^{N-1})$ to run the experiments.  Under a limited measurement budget $N$, the goal of the learner is to  recommend  an implementation decision $\bm{x}^{N}$ that maximizes $ p(y=+1|\bm{x}^{N})$. 

We adopt probabilistic models for classification. Under general assumptions, the probability of success can be written as a  link function acting on a linear function of the feature vector
$$
p(y=+1|\bm{x})=\sigma(\bm{w}^T\bm{x}).
$$
In this paper, we illustrate the ideas using the logistic link function $\sigma(a)=\frac{1}{1+\text{exp}(-a)}$ and probit link
function $\sigma(a)=\Phi(a)=\int_{-\infty}^a\mathcal{N}(s|0,1^2) \text{d}s$  given its analytic simplicity and popularity, but any monotonically
increasing function $\sigma: \mathbb{R} \mapsto [0,1]$ can be used. The main difference between the two sigmoid functions is that the logistic function has slightly heavier tails than the normal CDF. Classification using the logistic function is called logistic regression and that using the normal CDF is called probit regression. 

 We start with a multivariate  prior distribution  for the unknown parameter vector $\bm{w}$.  At iteration $n$, we choose an  alternative   $\bm{x}^{n}$ to measure and observe a binary outcome $y^{n+1}$ assuming labels are generated independently given $\bm{w}$. Each alternative can be evaluated more than once with potentially  different outcomes. Let $\mathcal{D}^n=\{(\bm{x}^{i},y^{i+1})\}_{i=0}^n$  denote the previous measured data set  for any $n=0,\dots,N$.  Define the filtration $(\mathcal{F}^n)_{n=0}^N$ by letting $\mathcal{F}^n$ be the sigma-algebra generated by $\bm{x}^0,y^1,\dots, \bm{x}^{n-1},y^n$. We  use $\mathcal{F}^n$ and $\mathcal{D}^n$ interchangeably.  Note that the notation here is slightly different from the (passive) PAC learning model  where the data  are i.i.d. and are denoted as $\{(\bm{x}_i, y_i)\}$. Yet in our (adaptive) sequential decision setting, measurement and implementation decisions $\bm{x}^{n}$ are restricted to be $\mathcal{F}^n$-measurable so that decisions
may only depend on measurements  made in the past. This notation with superscript indexing time stamp is standard, for example,  in control theory, stochastic optimization and optimal learning. We  use Bayes' theorem to form a sequence of posterior predictive distributions $  p(\bm{w}|\mathcal{D}^n)$. 

The next lemma states the equivalence of using true probabilities and sample estimates when evaluating a policy. The proof is left in the supplementary material.

\begin{lemma}\label{eqv}
Let $\Pi$ be the set of policies, $\pi \in \Pi$, and $\bm{x}^\pi = \arg \max_{\bm{x}}  p(y = +1 | \bm{x}, \mathcal{D}^N)$ be the implementation decision after the budget $N$ is exhausted. Then
$$
\mathbb{E}_{\bm{w}}[ p(y=+1|\bm{x}^\pi, \bm{w})]=\mathbb{E}_{\bm{w}}[\max_{\bm{x}} p(y=+1|\bm{x},\mathcal{D}^{N})],
$$
where the expectation is taken over the prior distribution of $\bm{w}$.
\end{lemma}

By denoting  $\mathcal{X}^I$ as an implementation policy for selecting an alternative after the measurement budget is exhausted, then $\mathcal{X}^I$ is a mapping from the history $\mathcal{D}^N$ to an alternative $\mathcal{X}^I(\mathcal{D}^N)$. Then as a corollary of Lemma \ref{eqv}, we have \cite{powell2012optimal}
$$\max_{\mathcal{X}^I}\mathbb{E}\big[  p\big(y = +1 | \mathcal{X}( \mathcal{D}^N)\big)\big]= \max_{\bm{x}} p(y = +1 | \bm{x}, \mathcal{D}^N).
$$
In other words, the optimal decision at time $N$ is to go with our final set of beliefs. By the equivalence of using true probabilities and sample estimates when evaluating a policy as stated in Lemma \ref{eqv}, while we want to learn the unknown true value $\max_{\bm{x}} p(y=+1|\bm{x})$, we may write our objective function  as
\begin{equation}\label{obj}
\max_{\pi \in \Pi} \mathbb{E}^{\pi}[ \max_{\bm{x}} p(y = +1 | \bm{x}, \mathcal{D}^N)].
\end{equation}

\section{Background: Linear classification}\label{BLR}
Linear classification, especially logistic regression, is  widely used in machine learning for binary classification \cite{hosmer2004applied}.  Assume that the probability of success $p(y=+1 | \bm{x})$ is a parameterized function $\sigma(\bm{w}^T\bm{x})$ and further assume that observations are independently of each other.  Given a training set $\mathcal{D}=\{(\bm{x}_i,y_i)\}_{i=1}^n$ with $\bm{x}_i$ a $d$-dimensional  vector and $y_i \in \{-1,+1\}$,  the likelihood function $ p(\mathcal{D}|\bm{w})$ is 
$
 p(\mathcal{D}|\bm{w}) = \prod_{i=1}^n \sigma(y_i\cdot\bm{w}^T\bm{x}_i).$
The weight vector $\bm{w}$ is found by maximizing the likelihood $ p(\mathcal{D}|\bm{w})$ or equivalently,
minimizing the negative log likelihood:
$$\min_{\bm{w}}\sum_{i=1}^n-\log(\sigma(y_i\cdot\bm{w}^T\bm{x}_i)).$$
In order to avoid over-fitting, especially when there are a large
number of parameters to be learned, $l_2$ regularization is often used. The estimate of the weight vector $\bm{w}$ is then  given by:
\begin{equation}\label{RLR}
\min_{\bm{w}} \frac{\lambda}{2}\|\bm{w}\|^2-\sum_{i=1}^n\log(\sigma(y_i\cdot\bm{w}^T\bm{x}_i)).
\end{equation}

 It can be shown that this log-likelihood function is globally concave in $\bm{w}$ for both logistic regression or probit regression.  As a result,  numerous optimization techniques are
available for solving it, such as steepest ascent, Newton's method and conjugate gradient ascent.

This logic is suitable for batch learning where we only need to conduct the minimization once to find the estimation of weight vector  $\bm{w}$ based on a given batch of training examples $\mathcal{D}$. Yet due to the sequential nature of our problem setting, observations come one by one as in online learning. After each new observation, if we retrain the linear classifier using all the previous data, we need to re-do the minimization, which is computationally inefficient. In this paper, we instead extend Bayesian  linear classification to perform  recursive  updates with each observation.

A Bayesian approach to linear classification models  requires a prior distribution for the weight parameters $\bm{w}$, and the ability to compute  the conditional posterior $p(\bm{w}| \mathcal{D})$ given the observation. Specifically, suppose we begin with an arbitrary prior $p(\bm{w})$
and apply Bayes' theorem to calculate the posterior:
$
p(\bm{w}|\mathcal{D}) =\frac{1}{Z} p(\mathcal{D}|\bm{w})p(\bm{w}),
$ where the normalization constant $Z$ is the unknown evidence.   An $l_2$-regularized logistic regression can be interpreted as a Bayesian model with a Gaussian prior on the weights with standard deviation $1/\sqrt{\lambda}$.

Unfortunately, exact Bayesian inference for linear classifiers is intractable since the evaluation of the posterior distribution comprises a product of sigmoid functions; in addition,  the integral in the normalization constant is intractable as well,  for both the  logistic function or probit function. We can either use analytic approximations to the posterior, or solutions based on Monte Carlo sampling, foregoing a closed-form expression for the posterior. In this paper, we consider different analytic approximations to the posterior to  make the computation tractable. 

\subsection{Online Bayesian probit regression based on assumed Gaussian density filtering}\label{sec:PR}
Assumed density filtering (ADF) is a general online learning schema for computing approximate posteriors in statistical models \cite{boyen1998tractable,lauritzen1992propagation,maybeck1982stochastic,sahami1998bayesian}. In ADF, observations are processed one by one, updating the posterior which is then approximated and is used as the prior distribution for  the next observation.

For a given  Gaussian prior distribution on some latent parameter $\bm{\theta}$, $p(\bm{\theta})= \mathcal{N}(\bm{\theta}|\bm{\mu},\bm{\Sigma})$ and a likelihood $t(\bm{\theta}):=p(\mathcal{D}|\bm{\theta})$, the posterior $p(\bm{\theta}|\mathcal{D})$ is generally non-Gaussian,
$$p(\bm{\theta}|\mathcal{D}) = \frac{t(\bm{\theta})p(\bm{\theta})}{\int t(\tilde{\bm{\theta}})p(\tilde{\bm{\theta}})\text{d}\tilde{\bm{\theta}}}.
$$
We find the best approximation by minimizing the Kullback-Leibler (KL) divergence between the true posterior $p(\bm{\theta}|\mathcal{D})$ and the Gaussian approximation. It is well known that when $q(x)$ is Gaussian, the distribution $q(x)$ that minimizes KL$(p(x)||q(x))$ is the one whose first and second moments match that of $p(x)$.  It can be shown that the  Gaussian approximation $\hat{q}(\bm{\theta})=\mathcal{N}(\bm{\theta}|\hat{\bm{\mu}},\hat{\bm{\Sigma}})$ found by moment matching is given as:
\begin{equation} \label{mm}
\hat{\bm{\mu}}=\bm{\mu}+\bm{\Sigma}\bm{g}, ~~~ \hat{\bm{\Sigma}}=\bm{\Sigma}-\bm{\Sigma}(\bm{g}\bm{g}^T-2\bm{G})\bm{\Sigma},
\end{equation}
where the vector $\bm{g}$ and the matrix $\bm{G}$ are given by
$$
\bm{g}=\frac{\partial \log \bigg( Z(\tilde{\bm{\mu}}, \tilde{\bm{\Sigma}})\bigg)}{\partial \tilde{\bm{\mu}}}|_{\tilde{\bm{\mu}} = \bm{\mu}, \tilde{\bm{\Sigma}}= \bm{\Sigma}}, ~~~\bm{G}=\frac{\partial \log \bigg( Z(\tilde{\bm{\mu}}, \tilde{\bm{\Sigma}})\bigg)}{\partial \tilde{\bm{\Sigma}}}|_{\tilde{\bm{\mu}} = \bm{\mu}, \tilde{\bm{\Sigma}}= \bm{\Sigma}},
$$
and the normalization function $Z$ is defined by
$$Z(\bm{\mu},\bm{\Sigma}) := \int t(\tilde{\bm{\theta}})\mathcal{N}(\tilde{\bm{\theta}}|\bm{\mu},\bm{\Sigma}) \text{d}\tilde{\bm{\theta}}. $$

For the sake of analytic convenience, we only consider probit regression  under assumed  Gaussian density filtering. Specifically, the distribution of $\bm{w}$ after $n$ observations is modeled as $p(\bm{w})=\mathcal{N}(\bm{w}| \bm{\mu}^n, \bm{\Sigma}^n)$. The likelihood function for the next available data $(\bm{x},y)$ is $t(\bm{w}):=\Phi(y\bm{w}^t\bm{x})$. Thus we have, $$p(\bm{w}|\bm{x},y) \propto \Phi(y\bm{w}^t\bm{x}) \mathcal{N}(\bm{w}|\bm{\mu}^n, \bm{\Sigma}^n).$$ Since the convolution of the normal CDF and a Gaussian distribution is another normal CDF,  moment matching \eqref{mm} results in an analytical solution to the Gaussian approximation:
\begin{eqnarray}
\bm{\mu}^{n+1} &=& \bm{\mu}^{n}+ \frac{y\bm{\Sigma}^{n}\bm{x}}{\sqrt{1+\bm{x}^T\bm{\Sigma}^n\bm{x}}}v \bigg(\frac{y\bm{x}^T\bm{\mu}^n}{\sqrt{1+\bm{x}^T\bm{\Sigma}^n\bm{x}}} \bigg), \\
\bm{\Sigma}^{n+1} &=& \bm{\Sigma}^n - \frac{(\bm{\Sigma}^n \bm{x})(\bm{\Sigma}^n \bm{x})^T}{1+\bm{x}^T\bm{\Sigma}^n\bm{x}}w \bigg(\frac{y\bm{x}^T\bm{\mu}^n}{\sqrt{1+\bm{x}^T\bm{\Sigma}^n\bm{x}}} \bigg),
\end{eqnarray}
where 
$$v(z):=\frac{\mathcal{N}(z|0,1)}{\Phi(z)} \text{ and } w(z):=v(z)\Big(v(z)+z\Big).$$

In this work, we focus on diagonal covariance matrices $\bm{\Sigma}^n$ with $(\sigma_i^{n})^2$ as the diagonal element due to computational simplicity and its equivalence with $l_2$ regularization, resulting in the  following update for the posterior parameters:
\begin{eqnarray}\label{op}
\mu^{n+1}_i &=& \mu^n_i + \frac{yx_i(\sigma_i^n)^2}{\tilde{\sigma}}v(\frac{y\bm{x}^T\bm{\mu}^n}{\tilde{\sigma}}), \\\label{opC}
(\sigma_i^{n+1})^2 &=& (\sigma_i^{n})^2 - \frac{x_i^2(\sigma^n_i)^4}{\tilde{\sigma}^2}w(\frac{y\bm{x}^T\bm{\mu}^n}{\tilde{\sigma}}),
\end{eqnarray}
where  $\tilde{\sigma}^2 :=1+\sum_{j=1}^d(\sigma_j^n)^2x_j^2$. See, for example, \cite{graepel2010web} and \cite{chu2011unbiased} for successful applications of this online probit regression model in prediction of click-through rates and stream-based active learning.

Due to the popularity of logistic regression and the computational limitations of ADF (on general link functions other than probit function), we develop an online Bayesian linear classification procedure for general link functions to recursively predict the response of each alternative in the next section.

\section{ Online Bayesian Linear Classification based on Laplace approximation}\label{sec:oblc}
In this section, we consider the Laplace approximation to the posterior and develop an online Bayesian linear classification schema for general link functions.    
\subsection{Laplace approximation}
Laplace's method aims to find a gaussian approximation to a probability density defined over a set of continuous variables.  It can be obtained by finding the mode of the posterior distribution and then fitting a Gaussian distribution centered at that mode \cite{bishop2006pattern}. Specifically, define the logarithm of the unnormalized posterior distribution as
 \begin{eqnarray}\nonumber 
 \Psi(\bm{w})=\log p(\mathcal{D}|\bm{w})+
\log p(\bm{w}). \end{eqnarray}
 Since the logarithm of a Gaussian distribution is a quadratic function, we consider a second-order Taylor expansion to $\Psi$ around its MAP (maximum a posteriori) solution $\hat{\bm{w}}= \arg \max_{\bm{w}}\Psi(\bm{w})$: 
 \begin{equation}\label{exp}
 \Psi(\bm{w}) \approx \Psi(\hat{\bm{w}})-\frac{1}{2}(\bm{w}-\hat{\bm{w}})^T \bm{H}(\bm{w}-\hat{\bm{w}}),
 \end{equation}
 where $\bm{H}$ is the Hessian of the negative log posterior evaluated at $\hat{\bm{w}}$:
 $$\bm{H}=-\nabla^2 \Psi(\bm{w})|_{\bm{w}=\hat{\bm{w}}}.$$
 By exponentiating both sides of Eq. \eqref{exp}, we can see that the Laplace approximation results in a normal approximation to the posterior 
\begin{equation}\label{pos}
p(\bm{w}|\mathcal{D}) \approx \mathcal{N}(\bm{w}|\hat{\bm{w}},\bm{H}^{-1}).
\end{equation}


For multivariate Gaussian priors $
p(\bm{w})= \mathcal{N}(\bm{w}|\bm{m}, \bm{\Sigma})
$,
\begin{equation}\label{LPD} 
\Psi(\bm{w}|\bm{m},\bm{\Sigma})=-\frac{1}{2}(\bm{w}-\bm{m})^T\bm{\Sigma}^{-1}(\bm{w}-\bm{m})+\sum_{i=1}^n\log(\sigma(y_i\cdot\bm{w}^T\bm{x}_i)), 
\end{equation}
 and the Hessian $\bm{H}$ evaluated at $\hat{\bm{w}}$ is given for both logistic and normal CDF link functions as:
\begin{equation}\label{hessian}
\bm{H}=\bm{\Sigma}^{-1}-\sum_{i=1}^n \hat{t}_i\bm{x}_i\bm{x}_i^T,
\end{equation} where $\hat{t}_{i}:= \frac{\partial^2 \log p(y_i|\bm{x}_i, \bm{w})}{\partial f_i^2}|_{f_i=\hat{\bm{w}}^T\bm{x}_i}$ and $f_i=\bm{w}^T\bm{x}_i$.

\subsection{Online Bayesian linear classification based on Laplace approximation}\label{sec:RBLR}
Starting from a Gaussian prior   $\mathcal{N}(\bm{w}|\bm{m}^0, \bm{\Sigma}^0)$,  after the first $n$ observations,  the Laplace approximated posterior distribution is $p(\bm{w}|\mathcal{D}^n) \approx \mathcal{N}(\bm{w}|\bm{m}^n, \bm{\Sigma}^n)$ according to \eqref{pos}. We formally define the state space $\mathcal{S}$ to be the cross-product of $\mathbb{R}^d$ and the space of positive semidefinite matrices. At each time $n$, our state of knowledge is thus $S^n=(\bm{m}^n, \bm{\Sigma}^n)$. Observations come one by one due to the sequential nature of our problem setting. After each new observation, if we retrain the Bayesian classifier using all the previous data, we need to calculate the MAP solution of \eqref{LPD} with $\mathcal{D} = \mathcal{D}^n$ to update from $S^n$ to $S^{n+1}$. It is computationally inefficient  even with a diagonal covariance matrix. It is better to extend the Bayesian linear classifier  to handle recursive  updates with each observation.

Here, we propose a fast and stable online algorithm for model updates with independent normal priors (with $ \bm{\Sigma}=\lambda^{-1} \bm{I}$, where $ \bm{I}$ is the identity matrix), which is equivalent to $l_2$ regularization and which also offers greater computational efficiency~\cite{wangKG2016}. At each time step $n$,   the Laplace approximated posterior  $\mathcal{N}(\bm{w}|\bm{m}^n, \bm{\Sigma}^n)$  serves as a prior to update the model when the next observation is made.  In this recursive way of model updating, previously measured data need not be stored or used for retraining the model. By setting the batch size $n=1$ in Eq. \eqref{LPD} and \eqref{hessian}, we have the sequential Bayesian linear model for classification as in Algorithm 1, where $\hat{t}:= \frac{\partial^2 \log( \sigma(yf))}{\partial f^2}|_{f=\hat{\bm{w}}^T\bm{x}}$.

\begin{algorithm}\label{RBLR}
\caption{Online Bayesian linear classification}
\SetKwInOut{Input}{input}\SetKwInOut{Output}{output}

 \Input{Regularization parameter $\lambda > 0$} 
 $m_j=0$,  $q_j=\lambda$. (Each weight $w_j$ has an independent prior $\mathcal{N}(w_j|m_j, q_j^{-1})$)\\
 \For{$t=1$ to $T$}{
 Get a new point $(\bm{x}, y)$.\\
Find $\hat{\bm{w}}$ as the maximizer of \eqref{LPD}: $-\frac{1}{2}\sum_{j=1}^d q_i(w_i-m_i)^2 +\log( \sigma(y_i\bm{w}^T\bm{x}_i)).$\\
$m_j=\hat{w}_j$\\
Update $q_i$ according to \eqref{hessian}:
$q_j \leftarrow q_j -\hat{t}x^2_{j}$.
}
\end{algorithm}

It is generally assumed that $\log \sigma(\cdot)$ is  concave to ensure a unique solution of Eq. \eqref{LPD}. It is satisfied by commonly used sigmoid functions for classification problems,
including logistic function, probit function, complementary log-log function $\sigma(a) = 1- \exp(-\exp(a))$  and log-log function $\exp(-\exp(-a))$. 

We can tap a wide range of convex optimization algorithms including gradient search, conjugate gradient, and BFGS method  \cite{wright1999numerical}. But if we set  $n=1$ and $\Sigma=\lambda^{-1} \bm{I}$ in Eq. \eqref{LPD}, a  stable and efficient algorithm for solving \begin{equation}\label{opE}  \arg\max_{\bm{w}}-\frac{1}{2}\sum_{j=1}^d q_i(w_i-m_i)^2 + \log(\sigma(y\bm{w}^T\bm{x})) \end{equation} can be obtained as follows. First, taking derivatives   with respect to $w_i$ and setting $\frac{\partial F}{\partial w_i}$ to zero, we have
\begin{equation*}
q_i(w_i-m_i)=\frac{yx_i\sigma'(y\bm{w}^Tx)}{\sigma(y\bm{w}^Tx)},~~~~i=1,2,\dots,d.\end{equation*}
Defining $p$ as
$$p:=\frac{\sigma'(y\bm{w}^Tx)}{\sigma(y\bm{w}^Tx)},$$
we then have $w_i=m_i+yp\frac{x_i}{q_i}.$ Plugging this back into the definition of $p$ to eliminate $w_i$'s, we get  the equation for $p$: $$p=\frac{\sigma'(p\sum_{i=1}^dx_i^2/q_i+y\bm{m}^Tx)}{\sigma(p\sum_{i=1}^dx_i^2/q_i+y\bm{m}^Tx)}.$$

 Since $\log(\sigma(\cdot))$ is concave, by its derivative we know the function $\sigma'/\sigma$ is monotonically decreasing, and thus the right hand side of the equation decreases as $p$ goes from $0$ to $\infty$.
We notice that  the right hand side  is positive when $p=0$ and
the left hand side is larger than the right hand side when $p=\sigma'(y\bm{m}^Tx)/\sigma(y\bm{m}^Tx)$. Hence the equation has a unique solution in interval $[0, \sigma'(y\bm{m}^Tx)/\sigma(y\bm{m}^Tx)]$.  A   simple one dimensional  bisection method is sufficient to efficiently find the root $p^* $ and thus the solution to the $d$-dimensional optimization problem \eqref{opE}.

We illustrate and validate this schema using   logistic and probit functions.  For logistic function $\sigma(\bm{w}^T\bm{x}) = - \log(1+\exp(-y\bm{w}^T\bm{x}))$, by setting 
$\partial F/\partial w_i=0$ for all $i$ and then by denoting $(1+\exp(y\mathbf{w}^T\mathbf{x}))^{-1}$ as $p$, we have
$$q_i(w_i-m_i)=ypx_i,~~~~i=1,2,\dots,d,$$ resulting in the following equation for $p$:
$$\frac{1}{p}=1+\exp\Big{(}y\sum_{i=1}^{d}(m_i+yp\frac{x_i}{q_i})x_i\Big{)}
=1+\exp(y\mathbf{m}^T\mathbf{x})\exp \Big{(}y^2p\sum_{i=1}^d\frac{x_i^2}{q_i}\Big{)}.$$

It is easy to see that the left hand side decreases from infinity to 1 and the right hand side increases from 1 when $p$ goes from 0 to 1, therefore the solution exists and is unique in $[0,1]$. 

For normal CDF link function $\sigma(\cdot)= \Phi(\cdot)$,  the computation is a little lengthier. First use  $\phi(\cdot)$ to denote the standard normal distribution $\mathcal{N}(\cdot|0,1^2)$, we have 
$$\frac{\partial F}{\partial w_i}= -q_i(w_i-m_i)+ \frac{yx_i\phi(\bm{w}^T\bm{x})}{\Phi(y\bm{w}^T\bm{x})}.$$ Let $p = \frac{\phi(\bm{w}^T\bm{x})}{\Phi(y\bm{w}^T\bm{x})}$ and $\partial F/\partial w_i=0$ for all $i$, and thus we have $w_i = m_i + yp\frac{x_i}{q_i}$. Plugging these into the definition of $p$ we have
\begin{equation}\label{eqq}
p= \frac{\phi\big(\sum_{i=1}^d (m_i + yp \frac{x_i}{q_i})x_i\big)}{\Phi\big(y\sum_{i=1}^d (m_i + yp \frac{x_i}{q_i})x_i\big)}.
\end{equation}
Define the right-hand-size of the above equation as $g(p)$. The next lemma shows that $\frac{\text{d}}{\text{d} { p}}  g(p) \le 0$ and thus the right-hand-side of Eq. \eqref{eqq} is non-increasing. Together with the fact that the left hand side increases from 0 to 1, the bisection method can also be used based on Eq. \eqref{eqq}. The proof can be found in Appendix  \ref{e3}.

\begin{lemma}\label{l2} Define $$g(p) :=\frac{\phi\big(\sum_{i=1}^d (m_i + yp \frac{x_i}{q_i})x_i\big)}{\Phi\big(y\sum_{i=1}^d (m_i + yp \frac{x_i}{q_i})x_i\big)}
.$$ We have $\frac{\text{d} }{\text{d} { p}} g(p) \le 0$  for every $p \in \mathbb{R}$.
\end{lemma}

\section{Knowledge Gradient Policy for Bayesian Linear  Classification Belief Model}\label{sec:KG}
We begin with a brief review of the knowledge gradient (KG) for ranking and selection problems, where each of the alternative can be measured sequentially to estimates its unknown underlying expected performance $\mu_x$. The goal is to   adaptively allocate alternatives to measure so as to  find an implementation decision that has the largest mean after the budget is exhausted. In a Bayesian setting, the performance of each alternative is represented by a (non-parametric) lookup table model of  Gaussian distribution. Specifically,  by imposing a Gaussian prior $\mathcal{N}(\bm{\mu}|\bm{\theta}^0, \bm{\Sigma}^0)$, the posterior after the first $n$ observations  is denoted by $\mathcal{N}(\bm{\mu}|\bm{\theta}^n, \bm{\Sigma}^n)$.  At the $n$th iteration, the knowledge gradient policy   chooses its ($n+1$)th measurement to maximize the single-period expected increase in value \cite{frazier2008knowledge}:
$$\nu_x^{\text{KG},n} = \mathbb{E}[\max_{x'}\theta_{x'}^{n+1}- \max_{x'}\theta_{x'}^{n}|x^n = x,S^n].$$
It enjoys nice properties, including myopic and asymptotic optimality. KG has been extended to various belief models (e.g.  \cite{mes2011hierarchical,negoescu2011knowledge,ryzhov2012knowledge,wang2015nested}). The knowledge gradient can  be extended to online problems where we need to maximize cumulative rewards \cite{ryzhov2012knowledge},
$$\nu_x^{\text{OLKG},n} =\theta_x^n + \tau \nu^{\text{KG},n}_x,$$
where $\tau$  reflects a planning horizon.

Yet there is no KG variant designed for binary classification with parametric models,  primarily because of the computational intractability  of dealing with nonlinear belief models. In what follows, we first formulate our learning problem as a Markov decision process and then extend the KG policy for stochastic binary outcomes where, for example,  each choice (say, a medical decision) influences the success or failure of a medical outcome.

\subsection{Markov decision process formulation} \label{MDP}
Our learning problem is a dynamic program that can be formulated as a Markov decision process. Define the state space $\mathcal{S}$ as the space of all possible predictive distributions for $\bm{w}$.  By Bayes' Theorem, the transition function $T$: $\mathcal{S}\times\mathcal{X}\times \{-1, 1\}$ is:
\begin{equation}\label{trf}
T\bigg(q(\bm{w}), \bm{x},y\bigg)\propto q(\bm{w})\sigma(y\bm{w}^T\bm{x}).
\end{equation}

 If we start from a Gaussian prior   $\mathcal{N}(\bm{w}|\bm{\mu}^0, \bm{\Sigma}^0)$,  after the first $n$ observed data,  the  approximated posterior distribution is $p(\bm{w}|\mathcal{D}^n) \approx \mathcal{N}(\bm{w}|\bm{\mu}^n, \bm{\Sigma}^n)$. The state space $\mathcal{S}$ is the cross-product of $\mathbb{R}^d$ and the space of positive semidefinite matrices.   The transition function for updating the belief state depends  on the belief model $\sigma(\cdot)$ and the approximation strategy. For example, for different update equations  in Algorithm 1 and  \eqref{op}\eqref{opC} under different approximation methods, the transition function can be defined as follows with degenerate state space  $\mathcal{S} := \mathbb{R}^d \times [0,\infty) ^d$:
 \begin{definition}The transition function based on online Bayesian classification with Laplace approximation $T$: $\mathcal{S}\times\mathcal{X}\times \{-1, 1\}$ is defined as
$$T^{\text{L}}\Big((\bm{\mu},\bm{\sigma}^2), \bm{x},y\Big) =\Big(\arg \min_{\bm{w}}\Psi \big(\bm{w}|\bm{\mu},\bm{\sigma}^{-2} \big),\bm{\sigma}^{-2}+ \hat{t}(y)\cdot \textbf{diag}(\bm{x}\bm{x}^T)\Big),
$$
where $\hat{t}(y):= \frac{\partial^2 \log p(y|\bm{x}, \bm{w})}{\partial f^2}|_{f=\hat{\bm{w}}^T\bm{x}}$ for either logistic or probit functions, $\textbf{diag}(\bm{x}\bm{x}^T)$ is a column vector containing the diagonal elements of $\bm{x}\bm{x}^T$ and $\bm{\sigma}^{-2}$ is understood as a column vector containing $\sigma_i^{-2}$, so that $S^{n+1}=T^{\text{L}}(S^n, \bm{x},Y^{n+1})$. $Y^{n+1}$ denotes the unobserved binary random variable at time $n$.
\end{definition}
\begin{definition}The transition function based on assumed density filtering $T$: $\mathcal{S}\times\mathcal{X}\times \{-1, 1\}$ is defined as
$$T^{\text{ADF}}\Big((\bm{\mu},\bm{\sigma}^2), \bm{x},y\Big) =\Big(\bm{\mu}+\frac{y\bm{x}^T\bm{\sigma}^2}{\tilde{\sigma}}v(\frac{y\bm{x}^T\bm{\mu}}{\tilde{\sigma}}),\bm{\sigma}^2- \frac{(\bm{x}^2)^T\bm{\sigma}^4}{\tilde{\sigma}^2}w(\frac{y\bm{x}^T\bm{\mu}}{\tilde{\sigma}}) \Big),
$$
where $\tilde{\sigma} = \sqrt{1+(\bm{x}^2)^T\bm{\sigma}^2}$, $v(z):=\frac{\mathcal{N}(z|0,1)}{\Phi(z)}$,  $w(z):=v(z)\Big(v(z)+z\Big)$ and $\bm{x}^2$ is understood as the column vector containing $x_i^2$, so that $S^{n+1}=T^{\text{ADF}}(S^n, \bm{x},Y^{n+1})$.
\end{definition}
In a dynamic program, the value function is defined as the value of the optimal policy given a particular state
$S^n$ at  time $n$, and may also be determined recursively through Bellman's equation.   In the case of stochastic binary feedback, the terminal value function $V^N: \mathcal{S} \mapsto \mathbb{R}$  is given by   \eqref{obj} as 
$$
V^{N}(s)=\max_{\bm{x}} p(y = +1 | \bm{x},s), \forall s \in \mathcal{S}.
$$

The dynamic programming principle tells us that the value function at any other time $n=1,\dots,N$, $V^n$, is given recursively by
$$
V^n(s)=\max_{\bm{x}}\mathbb{E}[V^{n+1}(T(s,\bm{x},Y^{n+1}))|\bm{x},s], \forall s \in \mathcal{S}.
$$

Since the curse of dimensionality on the state space $\mathcal{S}$
makes direct computation of the value function  intractable, computationally efficient approximate
policies need to be considered.  A computationally  attractive policy for ranking and selection problems is known as the knowledge
gradient (KG)  \cite{frazier2008knowledge}, which will be extended to handle Bayesian classification models in the next section.

\subsection{Knowledge Gradient for Binary Responses} \label{sec:offlineKG}
The knowledge gradient of measuring an alternative $\bm{x}$ can be defined as follows:
\begin{definition}The knowledge gradient of measuring an alternative $\bm{x}$ while in state $s$ is 
\begin{equation} \label{KG}
\nu_{\bm{x}}^{\text{KG}}(s) := \mathbb{E}\Big[ V^{N}\Big(T(s,\bm{x},Y)\Big)-V^{N}(s)|\bm{x},s \Big].
\end{equation}
\end{definition}

$V^{N}(s)$ is deterministic given $s$ and is independent of alternatives $\bm{x}$. Since the label for alternative $\bm{x}$ is not known at the time of selection,  the expectation is computed conditional on the current belief state  $s=(\bm{\mu},\bm{\Sigma})$. Specifically,
given a state $s=(\bm{\mu},\bm{\Sigma})$, the outcome $y$ of an alternative $\bm{x}$ is a random variable that follows from a Bernoulli  distribution with a predictive distribution 
\begin{eqnarray}\label{predictD}
 p(y=+1|\bm{x},s) =\int  p(y=+1|\bm{x},\bm{w}) p(\bm{w}|s)\text{d}\bm{w}
= \int \sigma(\bm{w}^T\bm{x})p(\bm{w}|s)\text{d}\bm{w}.
\end{eqnarray}

We can calculate the expected value in the next state as
\begin{eqnarray*}
&&\mathbb{E}[V^{N}(T(s,\bm{x},y))] \\
&=&  p(y=+1|\bm{x},s)V^N\Big(T(s, \bm{x},+1)\Big)+  p(y=-1|\bm{x},s)V^N\Big(T(s, \bm{x},-1)\Big)\\
&=& p(y=+1|\bm{x},s)\cdot  \max_{\bm{x}'} p\Big(y = +1 | \bm{x}',T(s,\bm{x},+1)\Big)\\
&&+ p(y=-1|\bm{x},s)\cdot  \max_{\bm{x}'} p\Big(y = +1 | \bm{x}',T(s,\bm{x},-1)\Big).
\end{eqnarray*}

The knowledge gradient policy suggests at each time $n$ selecting the alternative that maximizes $\nu_{\bm{x}}^{\text{KG},n}(s^{n})$ where  ties are broken randomly.  Because of the errors incurred by approximation and numerical calculation, the tie should be understood as within $\epsilon$-accuracy.  The knowledge gradient policy can work with any choice of link function $\sigma(\cdot)$ and approximation procedures by adjusting the transition function $T (s, x, \cdot)$ accordingly. That is, $T^{\text{L}}$ or $T^{\text{ADF}}$.

The predictive distribution $ \int \sigma(\bm{w}^T\bm{x})p(\bm{w}|s)\text{d}\bm{w}$ is obtained by marginalizing with respect to the distribution specified by  current belief state   $p(\bm{w}|s)=\mathcal{N}(\bm{w}|\bm{\mu},\bm{\Sigma})$. Denoting $a=\bm{w}^T\bm{x}$ and $\delta(\cdot)$ as the Dirac delta function, we have $\sigma(\bm{w}^T\bm{x})=\int \delta(a-\bm{w}^T\bm{x})\sigma(a)\text{d}a.$
Hence 
$$\int \sigma(\bm{w}^T\bm{x})p(\bm{w}|s) \text{d}\bm{w}=\int \sigma(a)p(a)\text{d}a,$$
where 
$p(a)=\int \delta(a-\bm{w}^T\bm{x})p(\bm{w}|s)  \text{d}\bm{w}.$
Since the delta function imposes a linear constraint on $\bm{w}$ and $p(\bm{w}|s) = \mathcal{N}(\bm{w}|\bm{\mu},\bm{\sigma}^2)$ is Gaussian, the marginal distribution $p(a)$ is also Gaussian. We can evaluate $p(a)$ by calculating the mean and variance of this distribution \cite{bishop2006pattern}. We have
\begin{eqnarray*}
\mu_a&=&\mathbb{E}[a]=\int p(a)a \text{ d}a = \int p(\bm{w}|s)\bm{w}^T\bm{x} \text{ d}\bm{w}=\bm{\mu}^T\bm{x},\\
\sigma_a^2&=& \text{Var}[a]=\int p(\bm{w}|s) \big((\bm{w}^T\bm{x})^2-(\bm{\mu}^T\bm{x})^2 \big) \text{ d}\bm{w}=\sum_{j=1}^d \sigma_j^2 x_j^2.
\end{eqnarray*} Thus $\int \sigma(\bm{w}^T\bm{x})p(\bm{w}|s) \text{d}\bm{w}=\int \sigma(a)p(a)\text{d}a=\int \sigma(a) \mathcal{N}(a|\mu_a, \sigma^2_a) \text{d}a.$

For probit function $\sigma(a)=\Phi(a)$, the convolution of  a Gaussian and a normal CDF can be evaluated analytically.  Thus for probit regression,  the predictive distribution can be solved exactly as 
$p(y= +1|\bm{x},s)= \Phi(\frac{ \mu_a}{\sqrt{1+\sigma^2_a}}).
$ Yet, the convolution of a Gaussian with a logistic sigmoid function  cannot be evaluated analytically.     We apply the approximation $\sigma(a) \approx \Phi(\alpha a)$ with $\alpha=\pi/8$  (see \cite{barber1998ensemble,spiegelhalter1990sequential}), leading to the following approximation for the convolution of a logistic sigmoid with a Gaussian 
$$
 p(y=+1|\bm{x},s)=\int \sigma(\bm{w}^T\bm{x})p(\bm{w}|s) \text{d}\bm{w} \approx \sigma(\kappa(\sigma^2_a)\mu_a),
$$ where $\kappa(\sigma^2)=(1+\pi \sigma^2/8)^{-1/2}$.

Because of the one-step look ahead, the KG calculation can  also benefit from the online recursive update of the belief either from ADF or from online Bayesian linear classification based on Laplace approximation. We summarize the decision rules of the knowledge gradient policy  at each iteration  under different sigmoid functions and different approximation methods in Algorithm 2, 3, and 4, respectively.

\begin{algorithm}\label{AlKG}
\caption{Knowledge Gradient Policy for Logistic Model based on Laplace approximation}
\SetKwInOut{Input}{input}\SetKwInOut{Output}{output}

 \Input{$m_j$, $q_j$  (Each weight $w_j$ has an independent prior $\mathcal{N}(w_j|m_j, q_j^{-1})$)}
 \For{$\bm{x}$ in $\mathcal{X}$}{
Let $\Psi(\bm{w},y)=-\frac{1}{2}\sum_{j=1}^d q_j(w_j-m_j)^2 - \log(1+\exp(-y\bm{w}^T\bm{x}))$\\
Use bisection  method to find 
$\hat{\bm{w}}_{+}=\arg \max_{\bm{w}}\Psi(\bm{w},+1), ~\hat{\bm{w}}_{-}=\arg \max_{\bm{w}}\Psi(\bm{w},-1)
$
$\mu=\bm{m}^T\bm{x}$, $\sigma^2=\sum_{j=1}^d q_j^{-1} x_j^2$\\
Let $\sigma(a)=\big(1+\exp(-a) \big)^{-1}$\\
 $\mu_{+}(\bm{x}'):=\hat{\bm{w}}_{+}^T\bm{x}'$,  $\mu_{-}(\bm{x}') :=\hat{\bm{w}}_{-}^T\bm{x}'$\\
$\sigma^2_{\pm}(\bm{x}'):= \sum_{j=1}^d\Big( q_j+\sigma( \hat{\bm{w}}^T_{\pm}\bm{x})(1-\sigma( \hat{\bm{w}}_{\pm}^T\bm{x}))x^2_{j} \Big)^{-1} (x'_j)^2$\\
$\tilde{\nu}_{\bm{x}}=\sigma(\kappa(\sigma^2)\mu)\cdot \max_{\bm{x'}}\sigma \big(\kappa(\sigma_+^2(\bm{x}'))\mu_{+}(\bm{x'})\big)+\sigma(-\kappa(\sigma^2)\mu)\cdot \max_{\bm{x'}}\sigma \big(\kappa(\sigma_-^2(\bm{x}'))\mu_{-}(\bm{x'})\big)$\\
}
$\bm{x}^{\text{KG}} \in \arg \max_{\bm{x}}\tilde{\nu}_{x}$
\end{algorithm}

\begin{algorithm}\label{KGP}
\caption{Knowledge Gradient Policy for Probit Model based on Laplace approximation}
\SetKwInOut{Input}{input}\SetKwInOut{Output}{output}

 \Input{$m_j$, $q_j$  (Each weight $w_j$ has an independent prior $\mathcal{N}(w_j|m_j, q_j^{-1})$)}
 
 \For{$\bm{x}$ in $\mathcal{X}$}{
Let $\Psi(\bm{w},y)=-\frac{1}{2}\sum_{j=1}^d q_i(w_i-m_i)^2 + \log(\Phi(y \bm{w}^T\bm{x}))$\\
Use bisection  method to find 
$\hat{\bm{w}}_{+}=\arg \max_{\bm{w}}\Psi(\bm{w},+1), \hat{\bm{w}}_{-}=\arg \max_{\bm{w}}\Psi(\bm{w},-1)$\\
$\mu=\bm{m}^T\bm{x}$, $\sigma^2=\sum_{j=1}^d q_j^{-1} x_j^2$\\
$\mu_{+}(\bm{x}') :=\hat{\bm{w}}_{+}^T\bm{x}'$,  $\mu_{-}(\bm{x}Õ):=\hat{\bm{w}}_{-}^T\bm{x}'$\\
$\sigma^2_{\pm}(\bm{x}') := \sum_{j=1}^d\Big( q_j+(\frac{\mathcal{N}(\hat{\bm{w}}^T_{\pm}\bm{x}|0,1)^2}{\Phi(\hat{\bm{w}}^T_{\pm}\bm{x})^2}+\frac{\hat{\bm{w}}^T_{\pm}\bm{x}\mathcal{N}(\hat{\bm{w}}^T_{\pm}\bm{x}|0,1)}{\Phi(\hat{\bm{w}}^T_{\pm}\bm{x})})x^2_{j} \Big)^{-1} (x'_j)^2$\\
$\tilde{\nu}_{\bm{x}}= \Phi(\frac{ \mu}{\sqrt{1+\sigma^2}})\cdot \max_{\bm{x'}}  \Phi(\frac{ \mu_+(\bm{x}')}{\sqrt{1+\sigma_+^2(\bm{x}')}})+\Phi(-\frac{ \mu}{\sqrt{1+\sigma^2}})\cdot \max_{\bm{x'}} \Phi(\frac{ \mu_-(\bm{x}')}{\sqrt{1+\sigma_-^2(\bm{x}')}})$\\
}
$\bm{x}^{\text{KG}} \in \arg \max_{\bm{x}}\tilde{\nu}_{x}$
\end{algorithm}

\begin{algorithm}\label{KGA}
\caption{Knowledge Gradient Policy for Probit Model based on assumed density filtering}
\SetKwInOut{Input}{input}\SetKwInOut{Output}{output}

 \Input{$m_j$, $q_j$  (Each weight $w_j$ has an independent prior $\mathcal{N}(w_j|m_j, q_j^{-1})$)}
 $\sigma_j^2=1/q_j$\\
 \For{$\bm{x}$ in $\mathcal{X}$}{
 Define $v(z):=\frac{\mathcal{N}(z|0,1)}{\Phi(z)} \text{ and } w(z):=v(z)\Big(v(z)+z\Big)$\\

$\mu=\bm{m}^T\bm{x}$, $\sigma^2=\sum_{j=1}^d \sigma_j^2 x_j^2$\\
$m_{+j} = m_{j} + \frac{x_j\sigma_j^2}{\sqrt{1+\sigma^2}}v(\frac{\bm{m}^T\bm{x}}{\sqrt{1+\sigma^2}})$, $m_{-j} = m_{j} - \frac{x_j\sigma_j^2}{\sqrt{1+\sigma^2}}v(-\frac{\bm{m}^T\bm{x}}{\sqrt{1+\sigma^2}})$\\
$\sigma_{+j}^2 = \sigma_{j}^2 - \frac{x_j^2\sigma_j^4}{\sqrt{1+\sigma^2}^2}w(\frac{\bm{m}^T\bm{x}}{\sqrt{1+\sigma^2}})$, $\sigma_{-j}^2 = \sigma_{j}^2 - \frac{x_j^2\sigma_j^4}{\sqrt{1+\sigma^2}^2}w(-\frac{\bm{m}^T\bm{x}}{\sqrt{1+\sigma^2}})$\\

$\mu_{+}(\bm{x}') :=\bm{m}_{+}^T\bm{x}'$,  $\mu_{-}(\bm{x}Õ):=\bm{m}_{-}^T\bm{x}'$\\
$\sigma_{+}^2(\bm{x}') := \sum_{j=1}^d  \sigma_{+j}^2(x'_j)^2$, $\sigma_{-}^2(\bm{x}') := \sum_{j=1}^d  \sigma_{-j}^2(x'_j)^2$\\
$\tilde{\nu}_{\bm{x}}= \Phi(\frac{ \mu}{\sqrt{1+\sigma^2}})\cdot \max_{\bm{x'}}  \Phi(\frac{ \mu_+(\bm{x}')}{\sqrt{1+\sigma_+^2(\bm{x}')}})+\Phi(-\frac{ \mu}{\sqrt{1+\sigma^2}})\cdot \max_{\bm{x'}} \Phi(\frac{ \mu_-(\bm{x}')}{\sqrt{1+\sigma_-^2(\bm{x}')}})$\\
}
$\bm{x}^{\text{KG}} \in \arg \max_{\bm{x}}\tilde{\nu}_{x}$
\end{algorithm}

We  next present the following finite-time bound on the the mean squared error (MSE)
of the estimated weight for Bayesian logistic regression. The proof is in the supplement. Since the learner plays an active role in selecting the measurements, the bound does not make the i.i.d. assumption of the examples which differs from the PAC learning bound. Without loss of generality, we assume $\|\bm{x}\|_2 \le 1$, $\forall{\bm{x}\in \mathcal{X}}$.  
In our finite-time analysis, we would like to highlight the difference between the ideal n-step estimate $\bm{w}^n=\arg\max_{\bm{w}}\Psi(\bm{w}|\bm{m}^0, \Sigma^0)$ with the prior distribution  $ p(\bm{w}^*)= \mathcal{N}(\bm{w}^*|\bm{m}^0, \bm{\Sigma}^0)$, and 
its approximation $\hat{\bm{w}}^n$. The approximation error may come from the Laplace approximation we adopt when an explicit form is not available,
or from accumulation of numerical error.

\begin{theorem}\label{finite}

Let $\mathcal{D}^n$ be the $n$ measurements produced by the KG policy.
Then with probability $P_d(M)$, the expected error of $\bm{w}^n$ is bounded as
$$\mathbb{E}_{ \bm{y}\sim \mathcal{B}(\mathcal{D}^n,\bm{w}^*)}||\bm{w}^n-\bm{w}^*||_2\le\frac{C_{min}+\lambda_{min}\big{(}\bm{\Sigma}^{-1}\big{)}}{2},$$
where the distribution  $\mathcal{B}(\mathcal{D}^n,\bm{w}^*)$ is the vector Bernoulli distribution
$Pr(y^i=+1)=\sigma((\bm{w}^*)^T\bm{x}^i)$,
$P_d(M)$ is the probability of a d-dimensional standard normal random variable appears in the ball with radius $M =\frac{1}{16}\frac{\lambda_{min}^2}{\sqrt{\lambda_{max}}}$ and
$C_{min}= \lambda_{min}
\Big{(}\frac{1}{n}\sum_{i=1} \sigma((\bm{w}^*)^T\bm{x}^i)\big{(}1-\sigma((\bm{w}^*)^T\bm{x}^i)\big{)}\bm{x}^i(\bm{x}^i)^T \Big{)}.$
The same finite time bound holds for $\hat{\bm{w}}^n$ as long as the approximation satisfies 
\begin{equation}\label{th1:con}
\Psi(\hat{\bm{w}}^n|\bm{m}^0, \Sigma^0) \ge \Psi(\bm{w}^n|\bm{m}^0, \Sigma^0) - \frac{1}{32}\lambda^3_{min}\big{(}\bm{\Sigma}^{-1}\big{)}.
\end{equation}
\end{theorem}
In the special case where $\bm{\Sigma}^0=\lambda^{-1}\bm{I}$, we have $\lambda_{max}=\lambda_{min}=\lambda$ and  $M=\frac{\lambda^{3/2}}{16}$.
The bound holds with higher probability  $P_d(M)$ with larger $\lambda$. This is natural since a larger $\lambda$ represents a normal distribution with narrower bandwidth, resulting in a more concentrated $\bm{w}^*$ around $\bm{m}^0$.

Condition \eqref{th1:con} quantifies how well the approximation should be carried out in order for the bound to hold.
It places a condition on the optimality of the log-likelihood value instead of the distance to the global maximizer.
This is a friendly condition since the dependence of the maximizer over the objective function is generally not continuous.
The approximation error could come from either the Laplace approximation or the numerical calculation in practice, 
which calls for experiments to further verity the performance of the algorithm.
Indeed, we conduct extensive experiments in Section \ref{EXP} to demonstrate the behaviors of the proposed algorithm together with bench mark methods.

If the goal of the learner is to maximize the cumulative successes as in bandit settings, rather than finding the final recommendation after the offline training phases, the online knowledge gradient policy can be modified by deleting $X^{\text{KG},n}$ at each time step $n$ as:
\begin{equation}\label{free}
X^{\text{KG},n}(S^n) = \arg \max_{\bm{x}}p(y=+1|x,S^n) + \tau\nu_{\bm{x}}^{\text{KG}}(S^n),
\end{equation}
where $\tau$ reflects a planning horizon.

\subsection{Behavior and Asymptotic Optimality}\label{5.3}
In this section, we study theoretically  the behavior of the KG policy, especially in the limit as the number of measurements $N$ grows large.  For the purposes of the theoretical analysis, we do not approximate the predictive posterior distribution. We use logistic function as the sigmoid link function
throughout this section. Yet the theoretical results can be generalized to other link functions.    We begin by showing the positive value of information (benefits of measurement).
\begin{proposition}[Benefits of measurement]\label{p1} The knowledge gradient of measuring any alternative $\bm{x}$ while in any state $s \in \mathcal{S}$ is nonnegative, $\mu_{\bm{x}}^{\text{KG}}(s) \ge 0.$ The state space $\mathcal{S}$ is the space of all possible predictive distributions for $\bm{w}$.
\end{proposition}

 The next lemma shows that the asymptotic probability of success of each alternative is well defined.
 \begin{lemma} 
For any alternative $\bm{x}$, $p_{\bm{x}}^n$ converges almost surely to a random variable $p_{\bm{x}}^\infty =\mathbb{E}\big[\sigma(\bm{w}^T\bm{x})|\mathcal{F}^\infty\big] $, where $p_{\bm{x}}^n$ is  short hand notation for $p(y=+1|\bm{x},\mathcal{F}^n)=\mathbb{E}\big[\sigma(\bm{w}^T\bm{x})|\mathcal{F}^n\big].$
\end{lemma}
\proof{Proof}Since $|\sigma(\bm{w}^T\bm{x})|\le 1$, the definition of $p_{\bm{x}}^n$ implies that  $p_{\bm{x}}^n$ is a bounded martingale and hence converges.
\endproof
The rest of this section shows that this limiting probability of success of each alternative is one in which the posterior is consistent and thus the KG is asymptotically optimal. We also show that as the number of measurements grows large, the maximum likelihood estimator  $\bm{w}_{\text{MLE}}$ based on the alternatives measured by the KG policy is consistent and asymptotically normal.   The next proposition states that if we have measured an alternative infinitely many times, there is no benefit to measuring it one more time. This is a key step for establishing the consistency of the KG policy and the MLE. The proof is similar to that by \cite{frazier2009knowledge} with additional mathematical steps for Bernoulli distributed random variables. See Appendix \ref{eaaa} for details.
\begin{proposition} \label{infty} If the policy $\pi$ measures alternative $\bm{x}$ infinitely often almost surely, then the value of information of that alternative $\nu_{\bm{x}}(\mathcal{F}^\infty) = 0$ almost surely under  policy $\pi$. 
\end{proposition}

Without loss of generality, we assume $\|\bm{x}\|_2 \le 1$ and  that the $d\times d$ matrix $P$ formed by $(\bm{x}_1,\bm{x}_2,\dots,\bm{x}_d)$ is invertible. The next theorem states the   strong consistency  and  asymptotic normality of the maximum likelihood estimator $\bm{w}^n_{\text{MLE}}$ (e.g. with $\lambda = 0$) based on KG's sequential selection of alternatives by verification of the following regularity conditions:
\begin{itemize}
    \item [(C$_1$)] The exogenous variables are uniformly bounded.
    \item [(C$_2$)] Let $\lambda_{1n}$ and $\lambda_{dn}$ be respectively the smallest and the largest eigenvalue of the Fisher information of the first $n$ observations $\bm{F}_n(\bm{w}^*)$. There exists $C$ such that $\lambda_{kN}/\lambda_{1N} < C$ , $\forall N$. 
    \item [(C$_3$)] The smallest eigenvalue of the Fisher information is divergent, $\lambda_{1N} \rightarrow +\infty.$
    \end{itemize}

\begin{theorem}\label{MLE} The sequence of $\bm{w}^n_{\text{MLE}}$ based on KG's sequential selection of alternatives converges almost surely to the true value $\bm{w}^*$ and is asymptotically normal:
$$\bm{F}_n^{\frac{1}{2}}(\bm{w}^n_{\text{MLE}} -  \bm{w}^*) \xrightarrow{d}\mathcal{N}(0,\bm{I}).$$ 
\end{theorem}
\proof{Proof}
We first prove that for any alternative $\bm{x}$, it will be measured infinitely many times almost surely. We will prove it by contradiction.
If this is not the case, then there exists a time $T$ such that for any $n>T$,
\begin{equation}\label{eq:consist1}
\mu^{\text{KG},n}_{\bm{x}}<\max_{\bm{x}\in \mathcal{X}} \mu^{\text{KG},n}_{\bm{x}}-\epsilon.
\end{equation}
This is because otherwise the difference between the KG value of $\bm{x}$ and the maximum KG value will be smaller than $\epsilon$ for infinitely many times, and thus
the probability of not measuring $\bm{x}$ after $T$ will be 0.

In addition, since the KG value is always non-negative, we have
$\max_{\bm{x}\in \mathcal{X}} \mu^{\text{KG},n}_{\bm{x}}>\epsilon$ for each $n>T$.
Notice that $\mathcal{X}$ is a finite set, then it follows that 
there exists an alternative $\bm{x}'$ such that the following happens infinitely many times:
\begin{equation}\label{eq:consist2}
\mu^{\text{KG},n}_{\bm{x}'}=\max_{\bm{x}\in \mathcal{X}} \mu^{\text{KG},n}_{\bm{x}}.
\end{equation}
Therefore, $\bm{x}'$ is measured infinitely many times almost surely.
However, we have proved in Proposition \ref{infty} that $\mu^{\text{KG},n}_{\bm{x}'}$ goes to 0 zero as long as we measure $\bm{x}'$ infinitely many times, which contradicts \eqref{eq:consist2}.
The contradiction shows that our original assumption that $\bm{x}$ only being measured finite times is incorrect.
As a consequence, we have prove that, almost surely, $\bm{x}$ will be measured infinitely many times.

Since our proof is for arbitrary $\bm{x}$, we actually proved that every alternative will be measured infinitely many times, which immediately leads to $\max_{\bm{x}\in \mathcal{X}} \mu^{\text{KG},n}_{\bm{x}}\rightarrow 0$.
Therefore the algorithm will eventually pick the alternative uniformly at random. Hence it satisfies the conditions (C$_2$) and (C$_3$), leading to the strong consistency and asymptotic normality \cite{gourieroux1981asymptotic, fahrmeir1985consistency,haberman1974analysis,cox1979theoretical}.

In particular, we can prove that the smallest eigenvalue of the Fisher's information goes
to infinity in a simple way. Without loss of generality, assume that the alternatives $\bm{x}_1,\bm{x}_2,\dots,\bm{x}_d$ are in general linear position, which means that the $d\times d$ matrix $P:=(\bm{x}_1,\bm{x}_2,\dots,\bm{x}_d)$ is invertible. We use $\mathcal{X}'$ to denote the set of these $d$ alternatives and denote $P^{-1}$ by $Q$.
Then $Q\bm{x}_t\bm{x}_t^TQ^T$ is a matrix whose every element equals 0 except that its $k$-th diagonal element equals one.

We use $\bm{F}_n$ to denote the Fisher's information at time $n$. 
Then from the fact that the matrix of type $\bm{x}\bm{x}^T$ is positive semi-definite,
 it is straightforward to see that 
$$\bm{F}_n(\bm{w})\ge \min_{\bm{x}\in\mathcal{X}}(1-\sigma(\bm{x}^T\bm{w}))\sigma(\bm{x}^T\bm{w})\sum_{1\le t\le n, \bm{x}_t \in \mathcal{X}'}\bm{x}_t\bm{x}_t^T,$$
where the constant  $\min_{\bm{x}\in\mathcal{X}}(1-\sigma(\bm{x}^T\bm{w}))\sigma(\bm{x}^T\bm{w})>0$ since $\mathcal{X}$ is a finite set.
Now define matrix $R_n$ as 
$$R_n:=\sum_{1\le t\le n, \bm{x}_t \in \mathcal{X}'}Q\bm{x}_t\bm{x}_t^TQ^T,$$
which is a diagonal matrix whose $i$-th element equals the times that $\bm{x}_i$ is estimated.
Since $\bm{x}_k$ is measured infinitely many times for $1\le k\le d$, 
then each diagonal element of $R_n$ goes to infinity.
Now notice that 
$\sum_{1\le t\le n, \bm{x}_t \in \mathcal{X}'}\bm{x}_t\bm{x}_t^T=Q^{-1}R_n(Q^{-1})^T$ is congruent to $R_n$ and $Q^{-1}$ is a constant matrix, then it follows that the smallest eigenvalue of $\sum_{1\le t\le n, \bm{x}_t \in \mathcal{X}'}\bm{x}_t\bm{x}_t^T$,
and hence the smallest eigenvalue of $\bm{F}_n$, goes to infinity. 

Similarly, by using the congruent argument, we only need to show that almost surely for any $k$,
there exists a constant $C$ such that $n_k(t)/t \ge C$, where $n_k(t)$ is the time that $n_k(t)$ is measured in the first $t$ measurements.
Since the algorithm eventually selects alternatives uniformly at random, without loss of generality, we assume $n_k(t)$ is the num of $t$ i.i.d Bernoulli random variable with probability $1/m$,
then it is clear to check that $\mathbb{E}[ n_k(t) ] = t/m$, and $Var[n_k(t)] = t(m-1)/m^2$.
Let $L$ be the event that $\underline{\lim}_{t\rightarrow\infty} n_k(t)/t < 1/2m$.
Notice that $L$ implies that there exists a sequence $t_1<t_2<t_3<\dots$ such that $n_k(t_s)/t_s < 1/2m$.
However, by Chebyshev's inequality, we have
$$\mathbb{P}\left[\frac{n_k(t_s)}{t_s} < \frac{1}{2m}\right] \le \mathbb{P}\left[\left | n_k(t_s) - \frac{t_s}{m}\right|\ge\frac{t_s}{2m}\right ] \le \frac{4(m-1)}{t_s}.$$
Notice that the above inequality holds for any $t_s$, which means $\mathbb{P}[L] = 0$ and thus $\underline{\lim}_{t\rightarrow\infty} n_k(t)/t \ge 1/2m $ almost surely,
which completes the proof.
\endproof

After establishing the consistency and asymptotic normality for $\bm{w}_{\text{MLE}}^n$, for any $\lambda>0$ as the inverse of the variance of the prior Gaussian distribution, the asymptotic bias of the estimator $\bm{w}^n_{\lambda}$ is as follows \cite{le1992ridge}:
$$\bm{E}[\bm{w}^n_{\lambda} - \bm{w}^*] = -2\lambda\{\bm{F}_n(\bm{w}^*) + 2 \lambda \bm{I}\}^{-1} \bm{w}^*,$$
and the asymptotic variance of $\bm{w}^n_{\lambda}$ is
$\{\bm{F}_n(\bm{w}^*) + 2 \lambda \bm{I}\}^{-1} \bm{F}_n(\bm{w}^*)  \{\bm{F}_n(\bm{w}^*) + 2 \lambda \bm{I}\}^{-1}.$

Finally, we show in the next theorem that given the opportunity to measure infinitely often, for any given neighborhood of $\bm{w}^*$, the probability that the posterior distribution lies in this neighborhood goes to 1 and  KG will discover
which one is the true best alternative. The detailed proof can be found in Appendix \ref{e5}.

\begin{theorem}[Asymptotic optimality]\label{consistency}
For any true parameter value $\bm{w}^* \in \mathbb{R}^d$ and any normal prior distribution with positive definite covariance matrix, under the knowledge gradient policy, the posterior is consistent and the KG policy is asymptotically optimal: $\arg\max_{\bm{x}}\mathbb{E}^{\text{KG}}[\sigma(\bm{w}^T\bm{x})|\mathcal{F}^\infty] =\arg\max_{\bm{x}} \sigma((\bm{w^*})^T\bm{x})$.

\end{theorem}

\subsection{Knowledge Gradient for contextual bandits} \label{context}

In  many of the applications, the ``success'' or ``failure'' can be predicted though an unknown relationship that depends on a \textbf{partially} controllable vector of attributes for each instance. For example, the patient attributes are given which we do not have control over. We can only choose a medical decision, and the outcome is based on  both the patient attributes and the selected medical decision. 

This problem is known as a contextual bandit \cite{zhangepoch,li2010contextual, krause2011contextual}. At each round $n$, the learner is presented with a  context vector $\bm{c}_n$ (patient attributes) and a set of actions $\bm{a} \in \mathcal{A}$ (medical decisions). After choosing an alternative $x$, we observe an outcome  $y^{n+1}$. The goal is to find a policy that selects actions such that the cumulative reward is as large as possible over time. In contrast, the previous section considers a {\it context-free} scenario with the goal of maximizing the probability of success after the offline training phase so that the error  incurred during the training is not punished.

Since there are very few patients with the same characteristics,  
it is unlikely that we can work with  one type of patients  that have the same characteristics to produce
statistically reliable performance measurements on medical outcomes. The contextual bandit formulation provides the potential to 
use parametric belief models that allow us to learn relationships across a wide range of patients
and medical decisions and personalize health care based on the patient attributes. 

Each of the past observations are made of triplets $(\bm{c}^n, \bm{a}^n,y^{n+1})$. Under Bayesian linear classification belief models, the binomial outcome $p(y=+1 | \bm{c},\bm{a} )$ is predicted though $\sigma \big(F(\bm{c},\bm{a})\big)$,
\begin{equation}\label{cm}
F(\bm{c},\bm{a}) = \theta + \bm{\alpha}^T\bm{\varphi_{\bm{c}}} +\bm{\beta}^T \bm{\psi_{\bm{a}}},
\end{equation}
 where suppose each context $\bm{c}$ and action $\bm{x}$ is represented by feature vectors $\bm{\varphi_{\bm{c}}}$ and  $\bm{\psi_{\bm{a}}}$, respectively. At each round $n$, the model updates can be slightly modified based on the  observation triplet $(\bm{c}^n, \bm{a}^n,y^{n+1})$ by treating $1|| \bm{\varphi_{\bm{c}}}|| \bm{\psi_{\bm{a}}}$ as the alternative $\bm{x}$, where $\bm{u}|| \bm{v}$ denotes the concatenation of the two vectors  $\bm{u}$ and $\bm{v}$.
 
The knowledge gradient  $\nu_{\bm{a}|\bm{c}}^{\text{KG}}(s)$ of measuring an action  $\bm{a}$ given context $\bm{c}$ can be defined as follows:
\begin{definition}The knowledge gradient of measuring an action $\bm{a}$  given  a context $\bm{c}$ while in state $s$ is 
\begin{equation} \label{KG}
\nu_{\bm{a}|\bm{c}}^{\text{KG}}(s) := \mathbb{E}\Big[ V^{N}\Big(T \big(s,1|| \bm{\varphi_{\bm{c}}}|| \bm{\psi_{\bm{a}}},y \big)\Big)-V^{N}(s)|\bm{c},\bm{a},s\Big].
\end{equation}
\end{definition}
The calculation of  $\nu_{\bm{a}|\bm{c}}^{\text{KG}}(s)$ can be modified based on Section \ref{sec:offlineKG} by replacing $\bm{x}$ as $1|| \bm{\varphi_{\bm{c}}}|| \bm{\psi_{\bm{a}}}$ throughout.

Since the objective in the contextual bandit problems is to maximize the cumulative number of successes, the knowledge gradient policy developed in Section \ref{sec:offlineKG} for stochastic binary feedback can be easily extended to online learning  following the  ``stop-learning'' (SL) policy adopted by \cite{ryzhov2012knowledge}. The action $X^{\text{KG},n}_{\bm{c}} (s^n)$ that is chosen by KG at time $n$ given a context $\bm{c}$ and a knowledge state $s^n$ can be obtained as:
$$X^{\text{KG},n}_{\bm{c}}(S^n) = \arg \max_{\bm{a}} p(y=+1| \bm{c}, \bm{a}, S^n) + \tau \nu_{\bm{a}|\bm{c}}^{\text{KG},n}(S^n),$$ where $\tau$ reflects a planning horizon.

It should be noted that if the context are constant over time, $X^{\text{KG},n}_{\bm{c}}(S^n)$ is degenerated to the case of {\it context-free} multi-armed bandit problems as discussed in Eq. \eqref{free}.

\section{Experimental results} \label{EXP}
In this section, we evaluate the proposed method in offline learning settings where we are not punished
for errors incurred during training and only concern with
the final recommendation after the offline training phases. 

We experiment with both synthetic datasets and the UCI machine learning repository  \cite{Lichman:2013} which includes classification problems drawn from settings including sonar, glass identification, blood transfusion, survival, breast cancer  (wpbc), planning relax and climate model  failure. We first  analyze the behavior of the KG policy and then compare it to state-of-the-art  learning algorithms. On synthetic datasets, we randomly generate a set of $M$ $d$-dimensional alternatives  $\bm{x}$ from $[-3,3]$. At each run, the stochastic binary labels are simulated using a $d+1$-dimensional weight vector $\bm{w}^*$ which is  sampled from the prior distribution $w^*_i \sim \mathcal{N}(0, \lambda)$. The $+1$ label for each alternative $\bm{x}$ is generated with probability $\sigma(w^*_0+\sum_{j=1}^d w^*_dx_d)$. For each  UCI dataset, we use all the data points as the set of alternatives with their original attributes. We then simulate their labels using a weight vector $\bm{w}^*$. This weight vector could have been chosen arbitrarily, but it was in fact a perturbed version of the weight vector trained through logistic regression on the original dataset. 

\subsection{Behavior of the KG policy}
\begin{figure}
    \centering
    \hspace*{-0.4cm}
    \begin{tabular}{cccc}
            \includegraphics[width=0.23\textwidth]{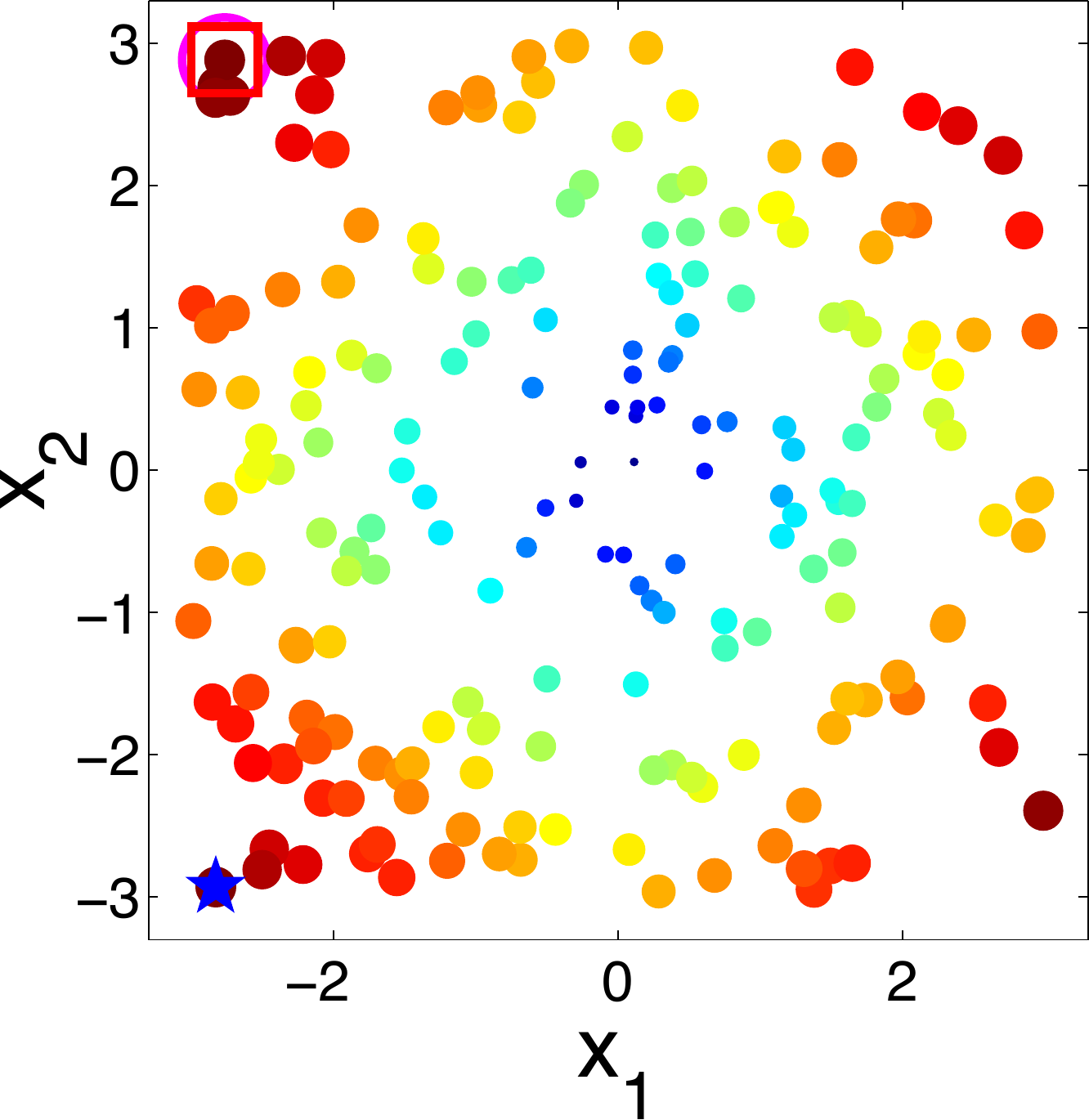}
  \includegraphics[width=0.23\textwidth]{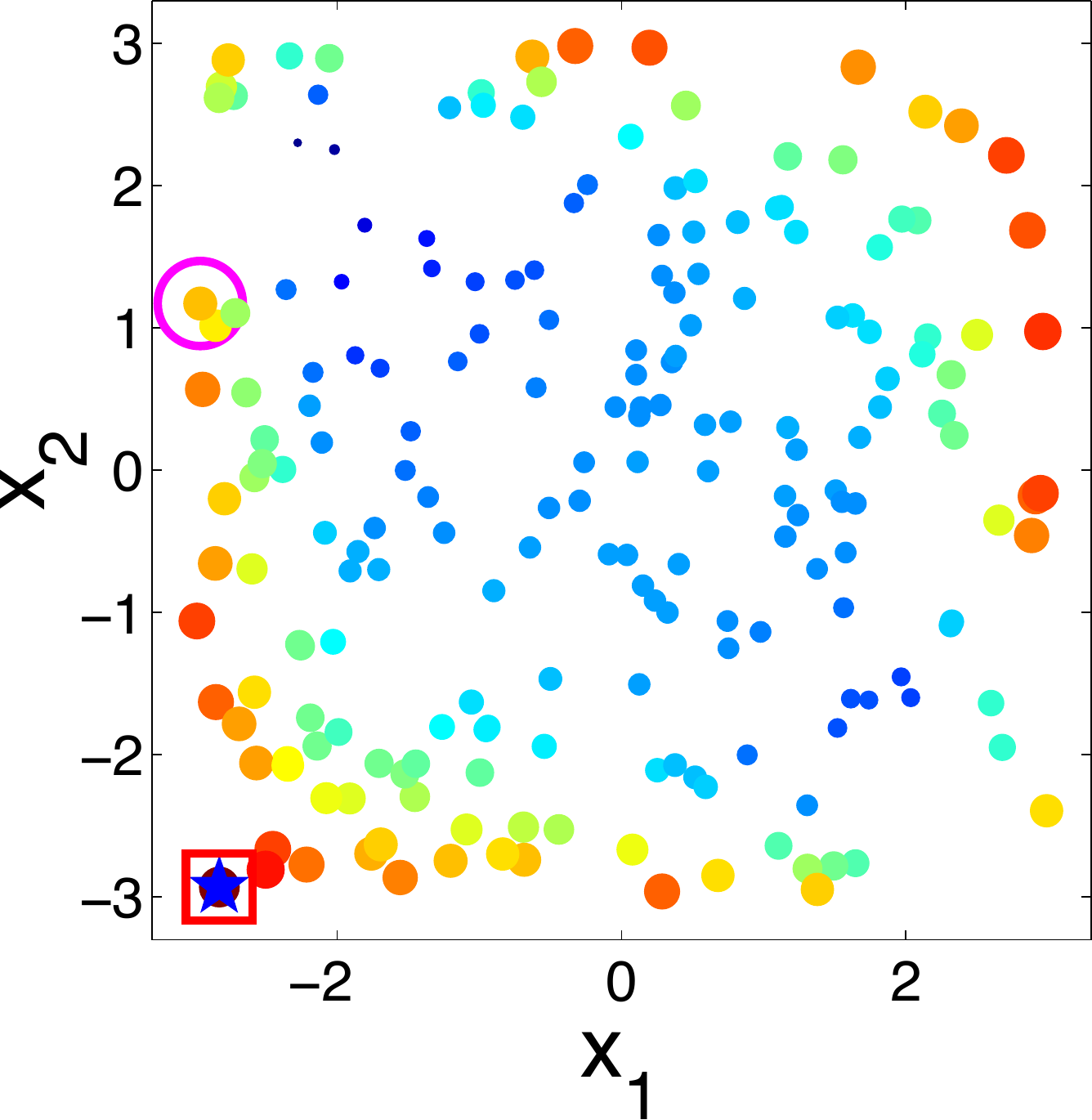} 
 \includegraphics[width=0.23\textwidth]{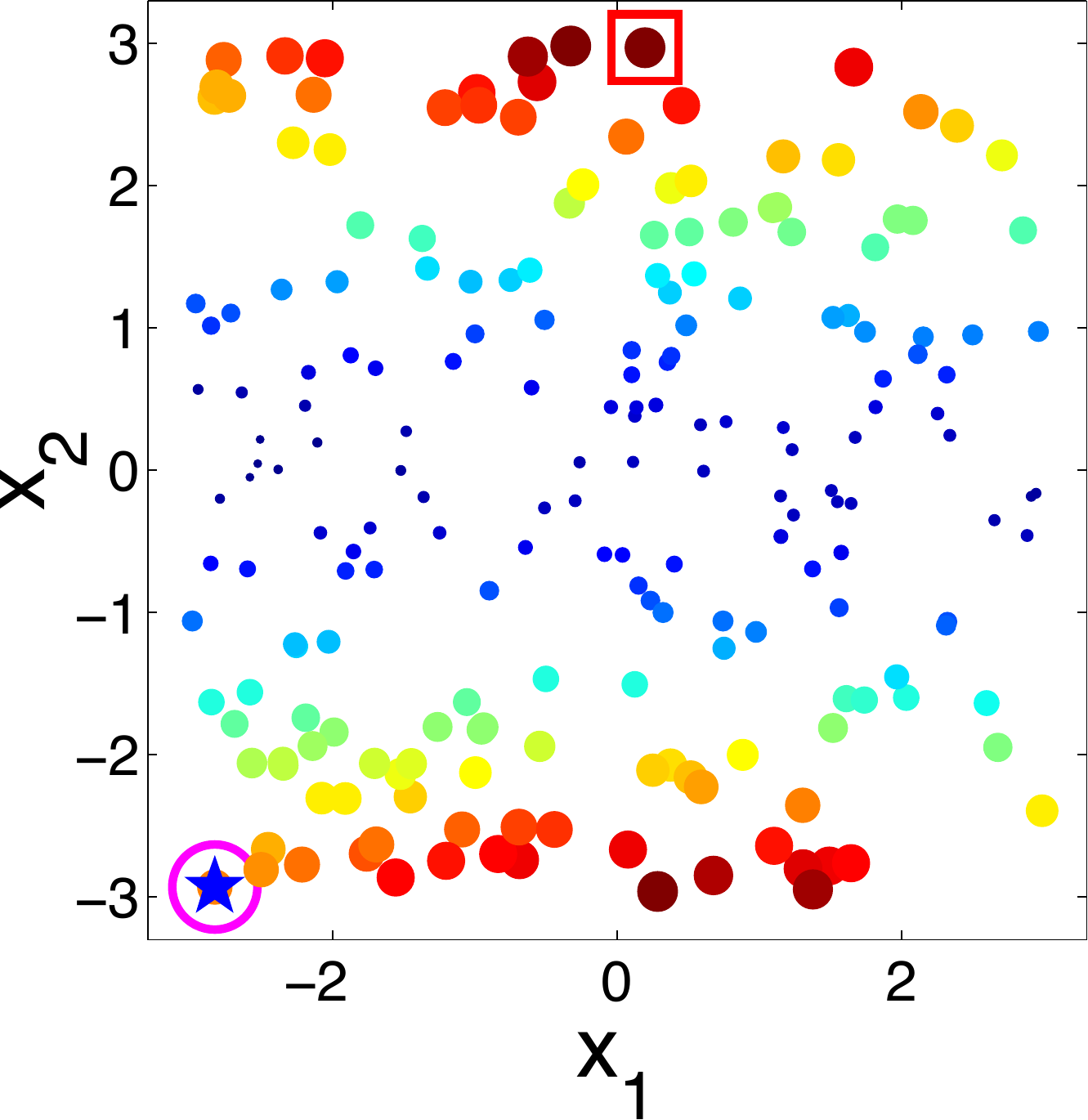} 
      \includegraphics[width=0.23\textwidth]{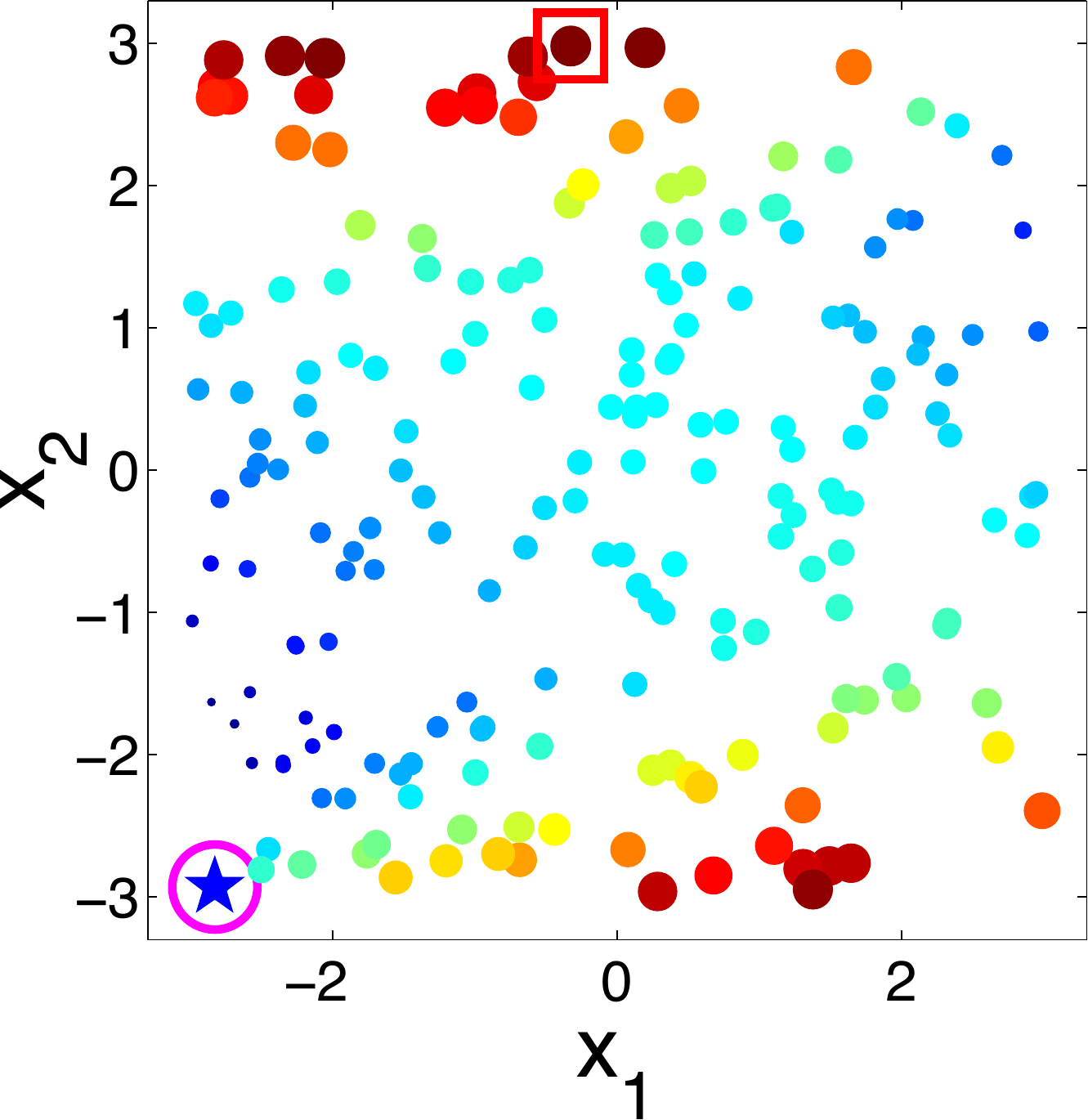}

    \end{tabular}
    \caption{The scatter plots illustrate the KG values at 1-4 iterations from left to right with both the color and the size  reflecting the magnitude. The star, the red square and pink circle indicate the true best alternative, the alternative to be selected and  the implementation decision, respectively. \label{2d}}
\end{figure}

To better understand the behavior of the KG policy,  Fig. \ref{2d} shows the snapshot of the KG policy at each iteration on a $2$-dimensional  synthetic dataset and a $3$-dimensional synthetic dataset in one run. Fig. \ref{2d} shows the snapshot on a 2-dimensional dataset with 200 alternatives. The scatter plots show the KG values with both the color and the size of the point reflecting the KG value of the corresponding alternative. The star denotes the true alternative with the largest response. The red square is the alternative with the largest KG value. The pink circle is the implementation decision that maximizes the response under current estimation of $\bm{w}^*$ if the budget is exhausted after that iteration.

It can be seen from the figure that the KG policy finds the true best alternative after only three measurements, reaching out to different alternatives to improve its estimates.  We can infer from Fig. \ref{2d} that the KG policy tends to choose alternatives near the boundary of the region.
This criterion is natural since in order to find the true maximum, we need to get enough information about $\bm{w}^*$ and 
estimate well the probability of points near the true maximum which appears near the boundary. On the other hand,  in a logistic model with labeling noise, a data $\bm{x}$ with small $\bm{x}^T\bm{x}$ inherently brings  little  information as pointed out by \cite{zhang2000value}. For an extreme example, when $\bm{x}=\bm{0}$ the label is always completely random for any $\bm{w}$ since $ p(y=+1|\bm{w},\bm{0}) \equiv 0.5$.  This is an issue when perfect classification is not achievable. So it is essential to label a data with larger $\bm{x}^T\bm{x}$ that has the most potential to  enhance its confidence non-randomly. 

\begin{figure}
    \hspace*{-0.5cm}
    \centering
    \begin{tabular}{ccc}
       \includegraphics[width=0.2\textwidth]{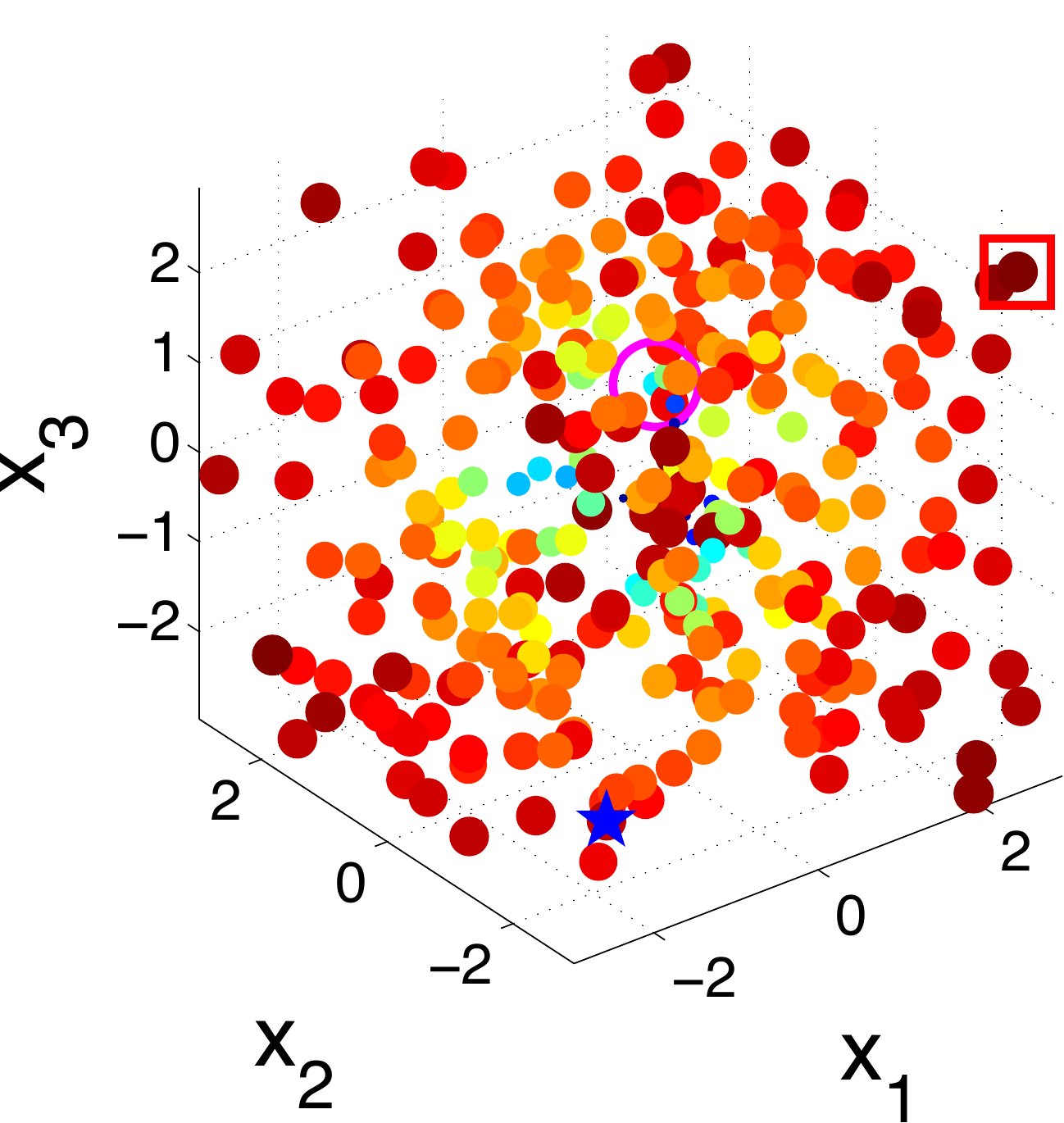}
 \includegraphics[width=0.2\textwidth]{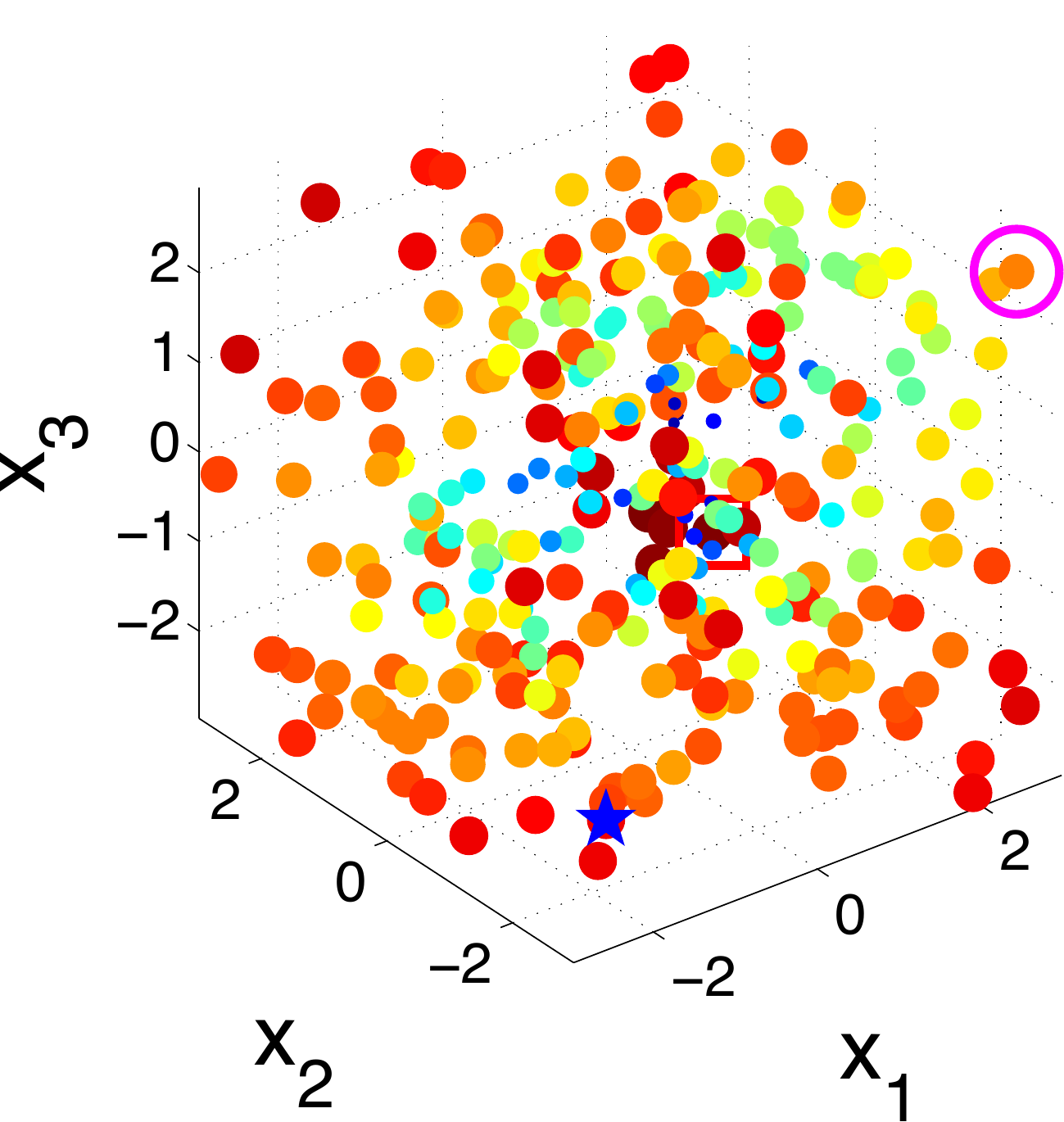} 
 \includegraphics[width=0.2\textwidth]{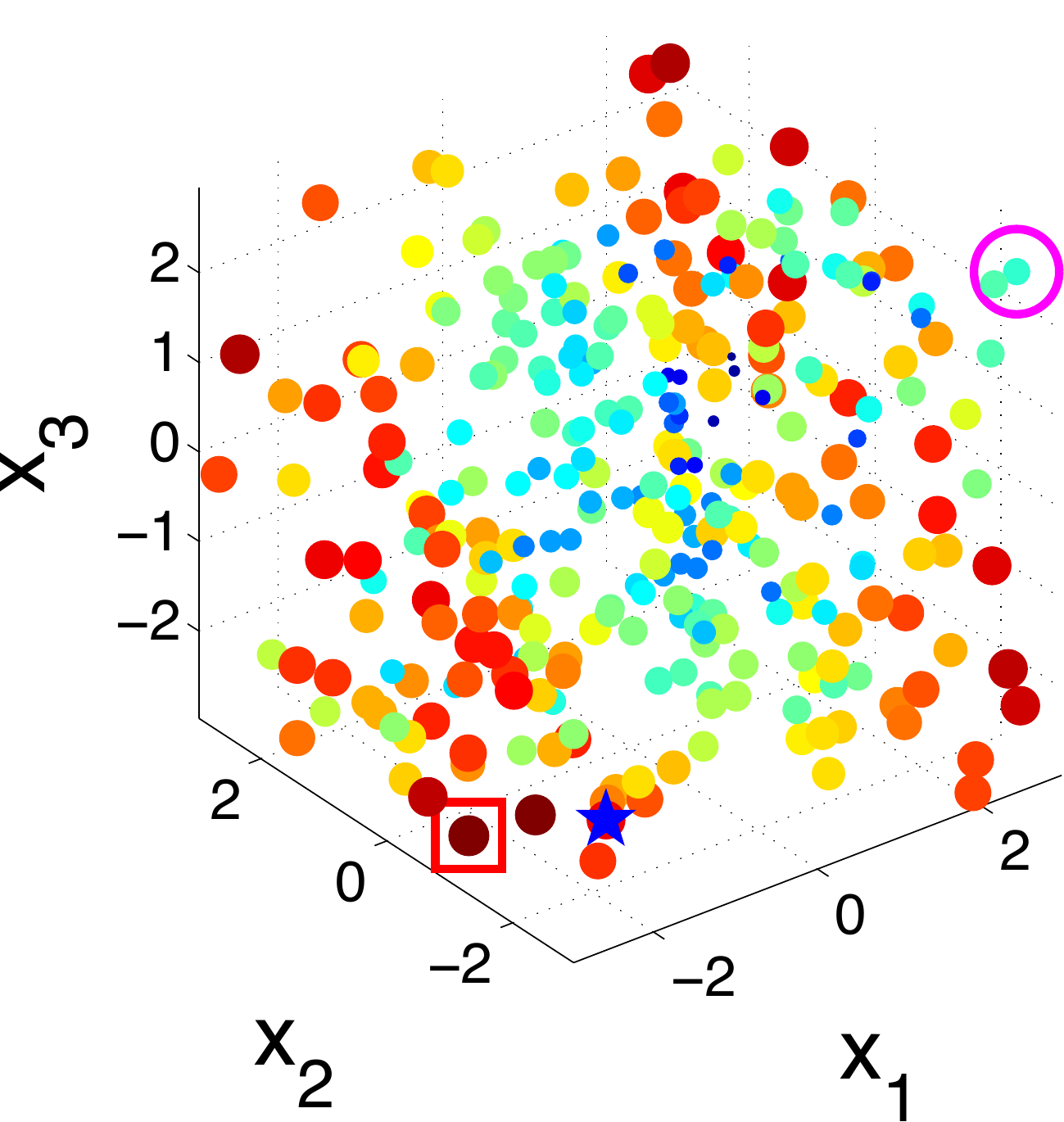} 
        \includegraphics[width=0.2\textwidth]{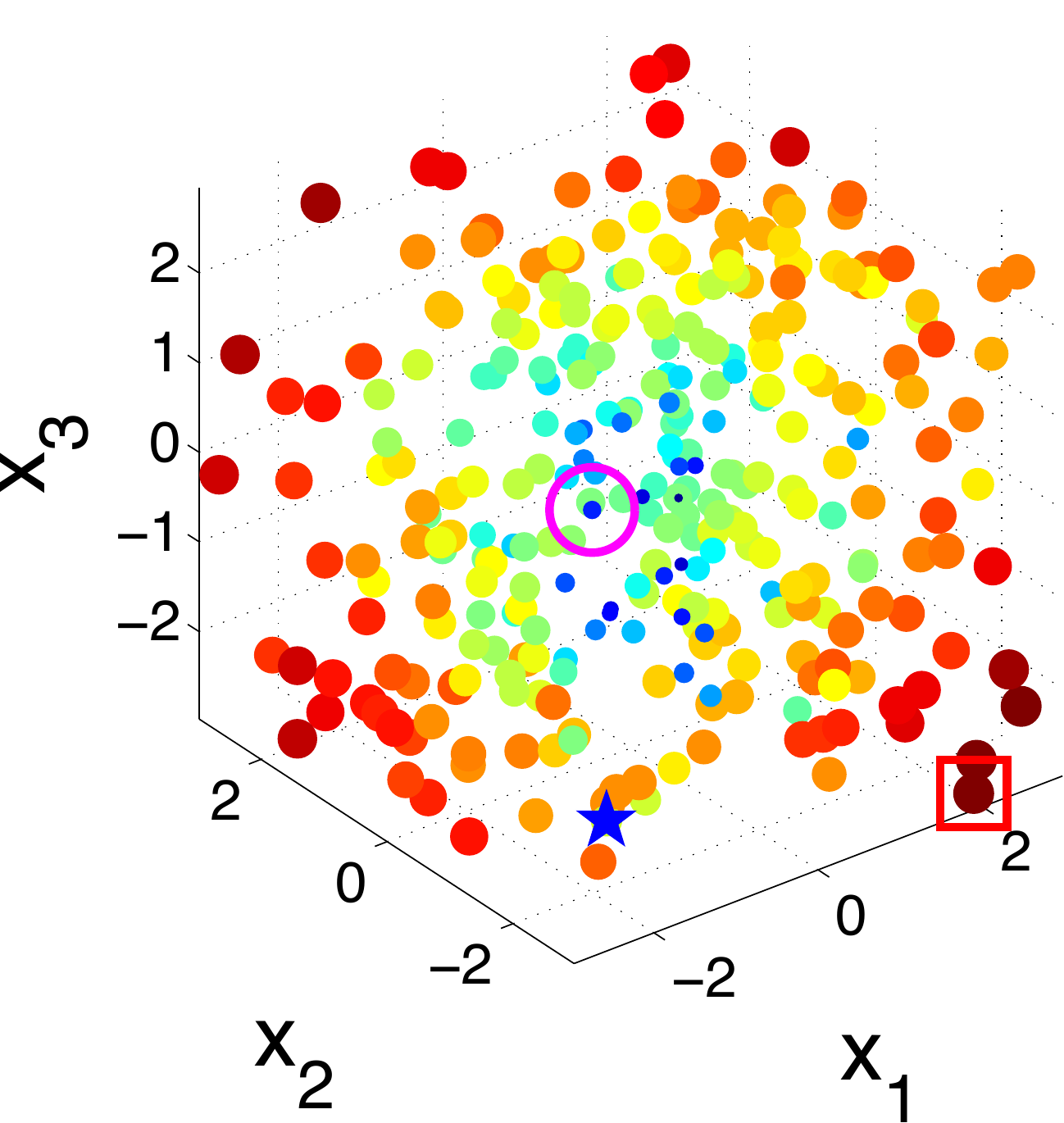}
         \includegraphics[width=0.2\textwidth]{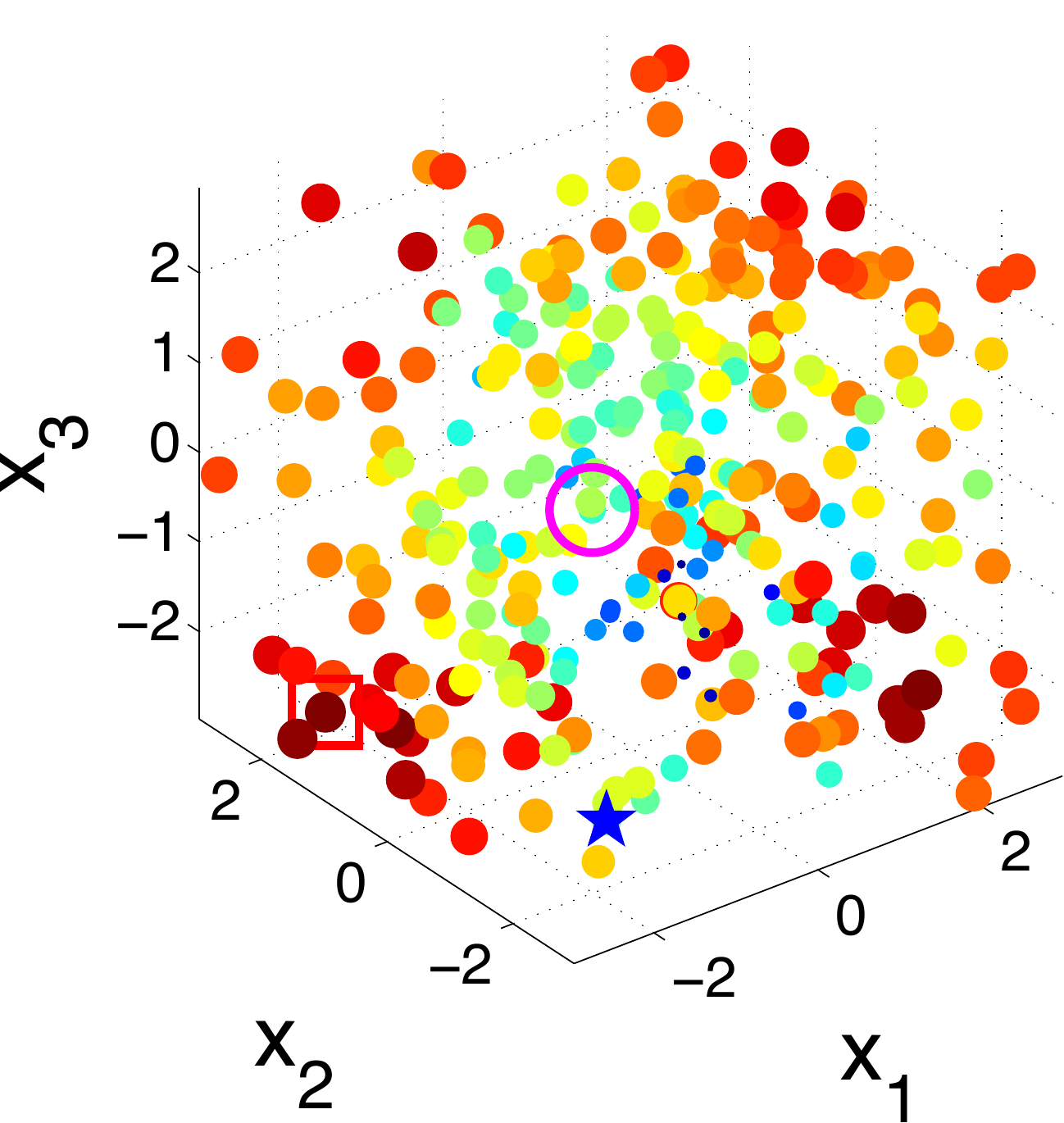} \\
 \includegraphics[width=0.2\textwidth]{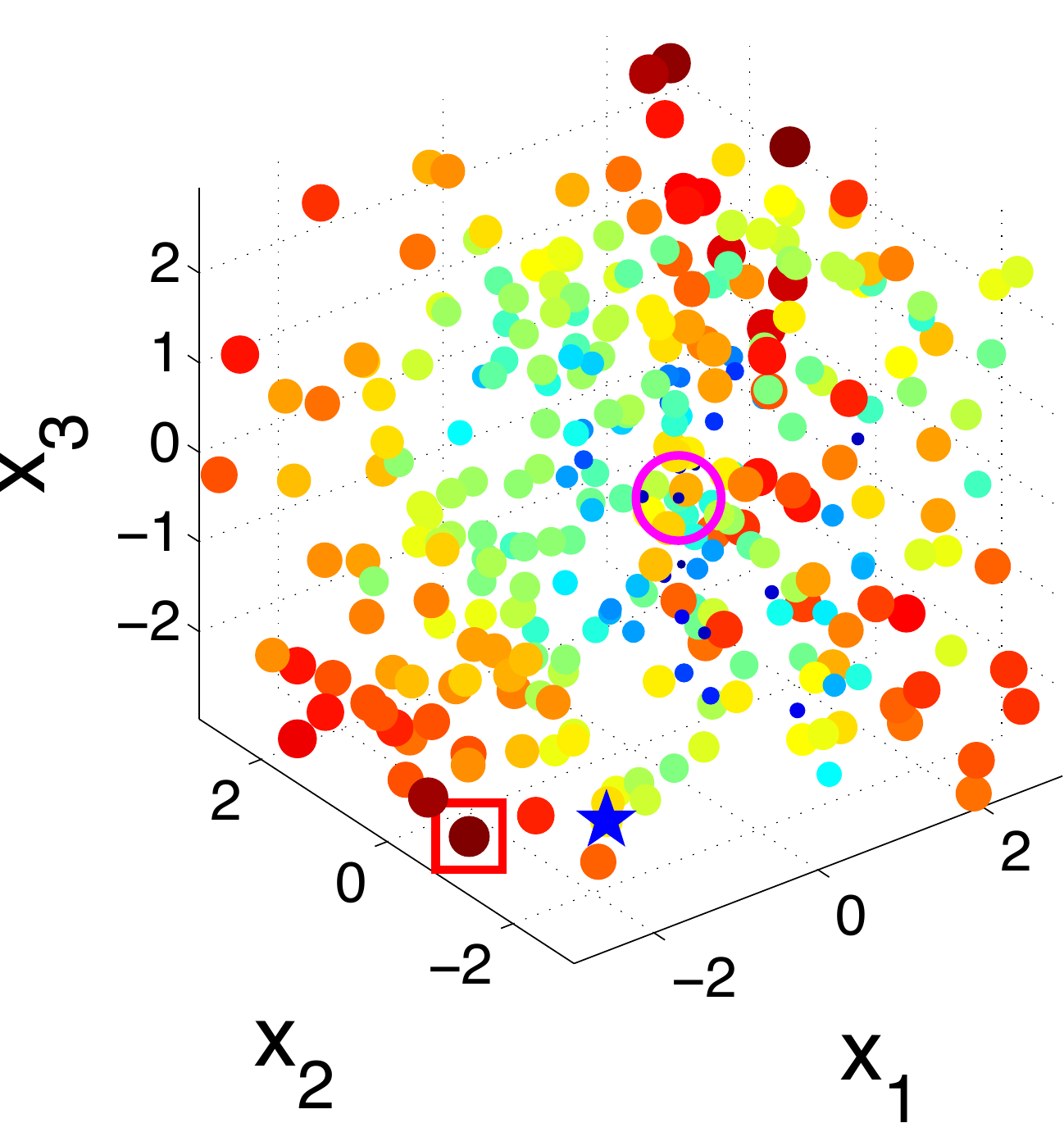} 
 \includegraphics[width=0.2\textwidth]{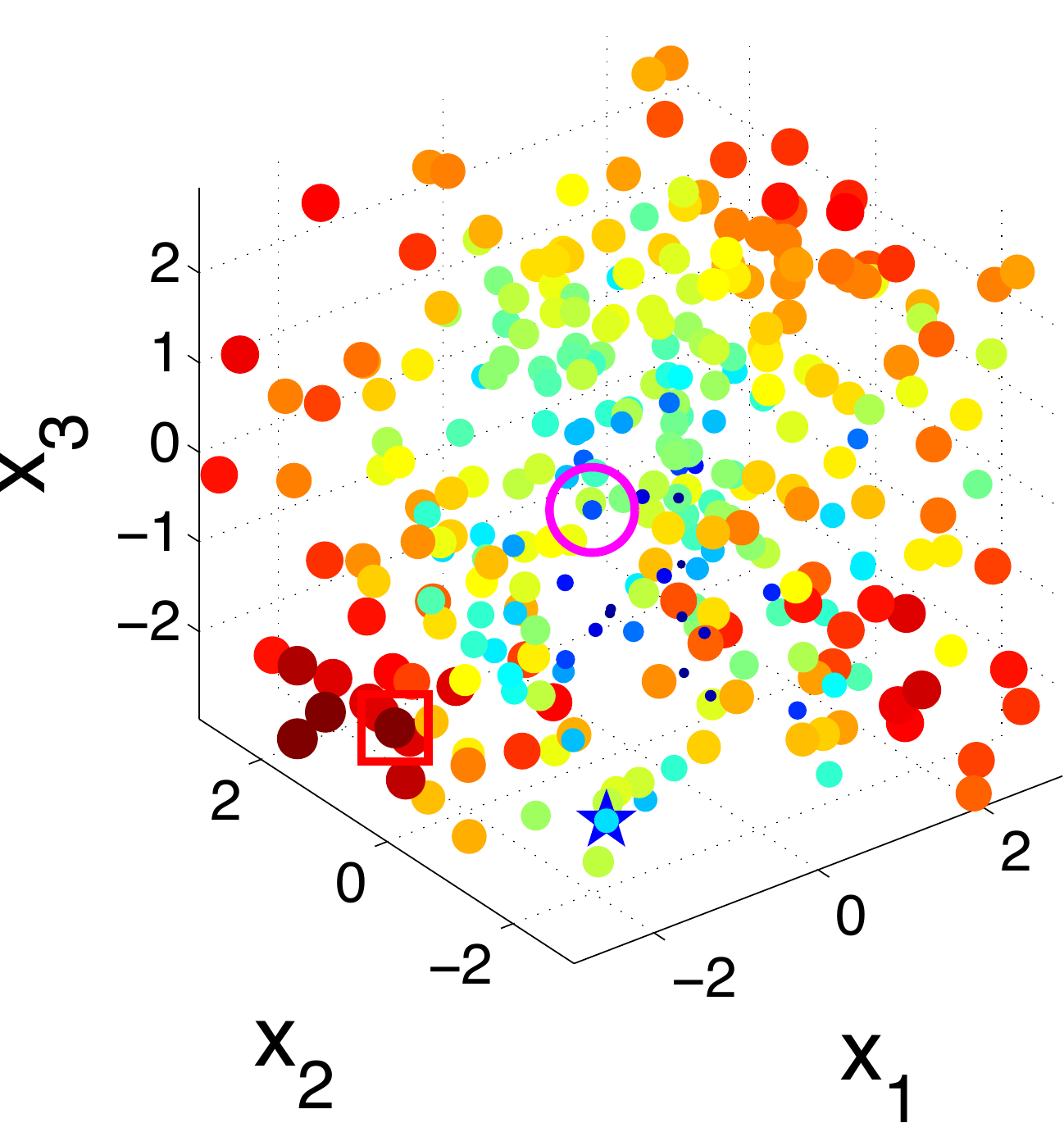}       
   \includegraphics[width=0.2\textwidth]{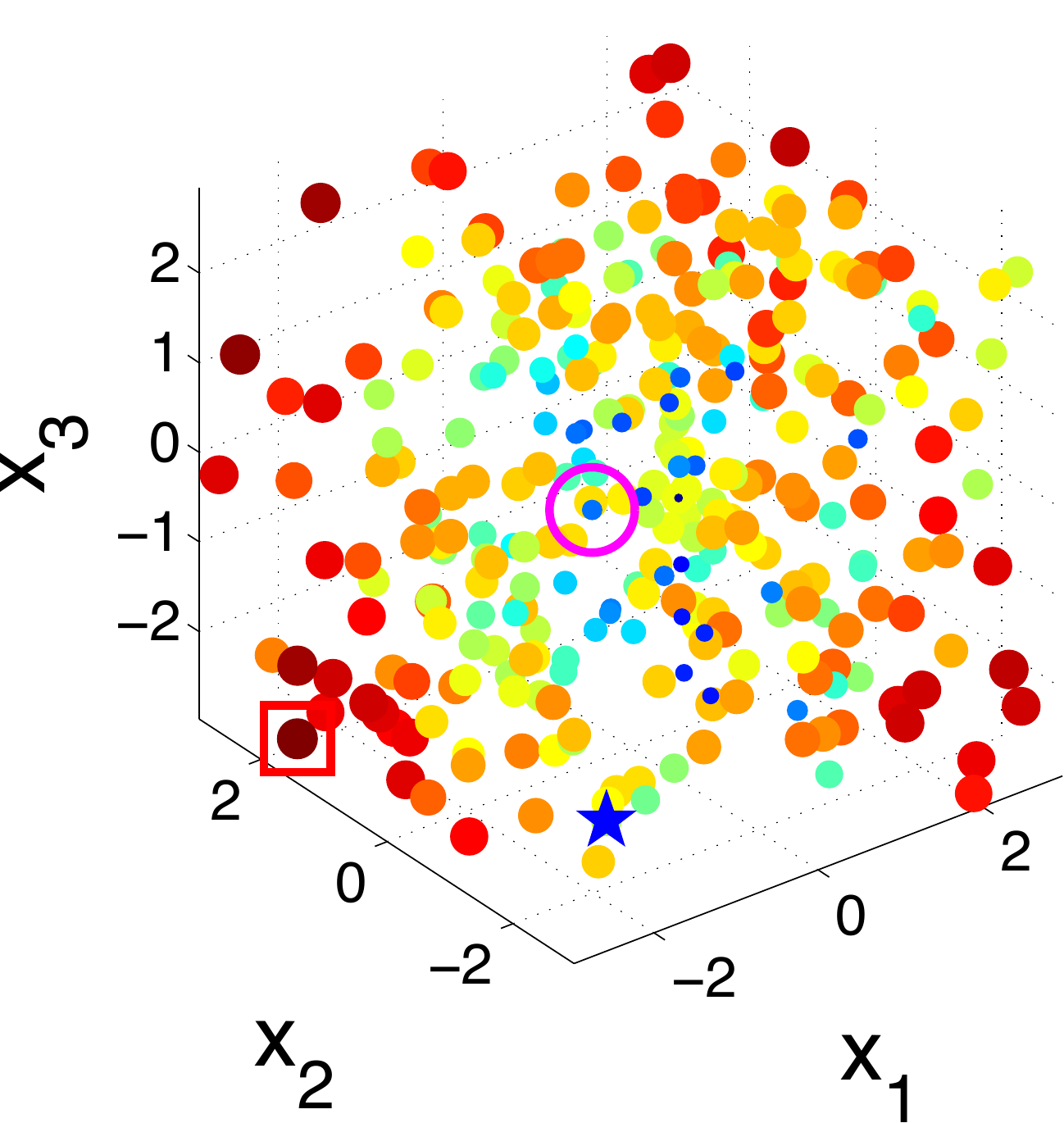}
 \includegraphics[width=0.2\textwidth]{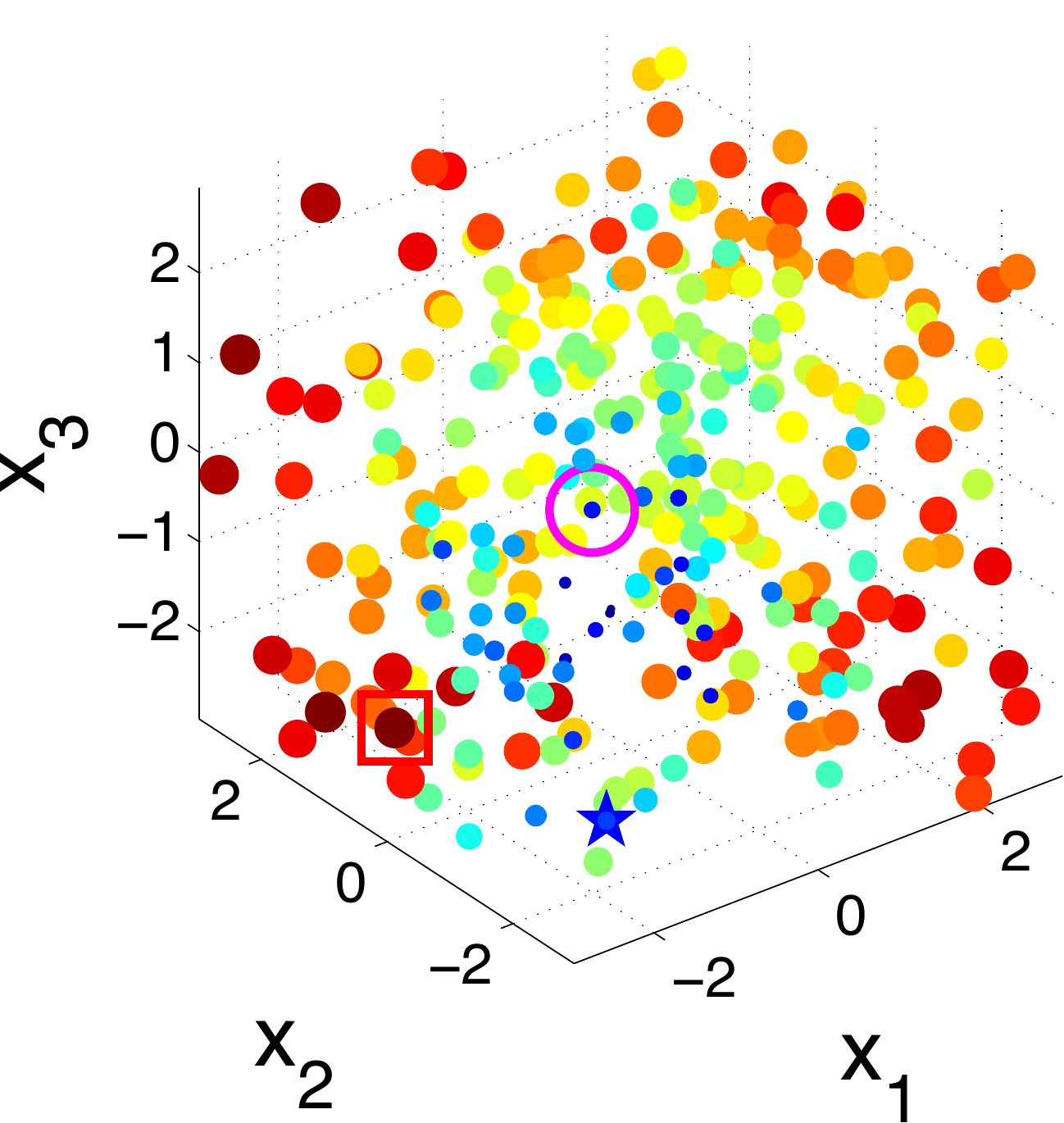} 
  \includegraphics[width=0.2\textwidth]{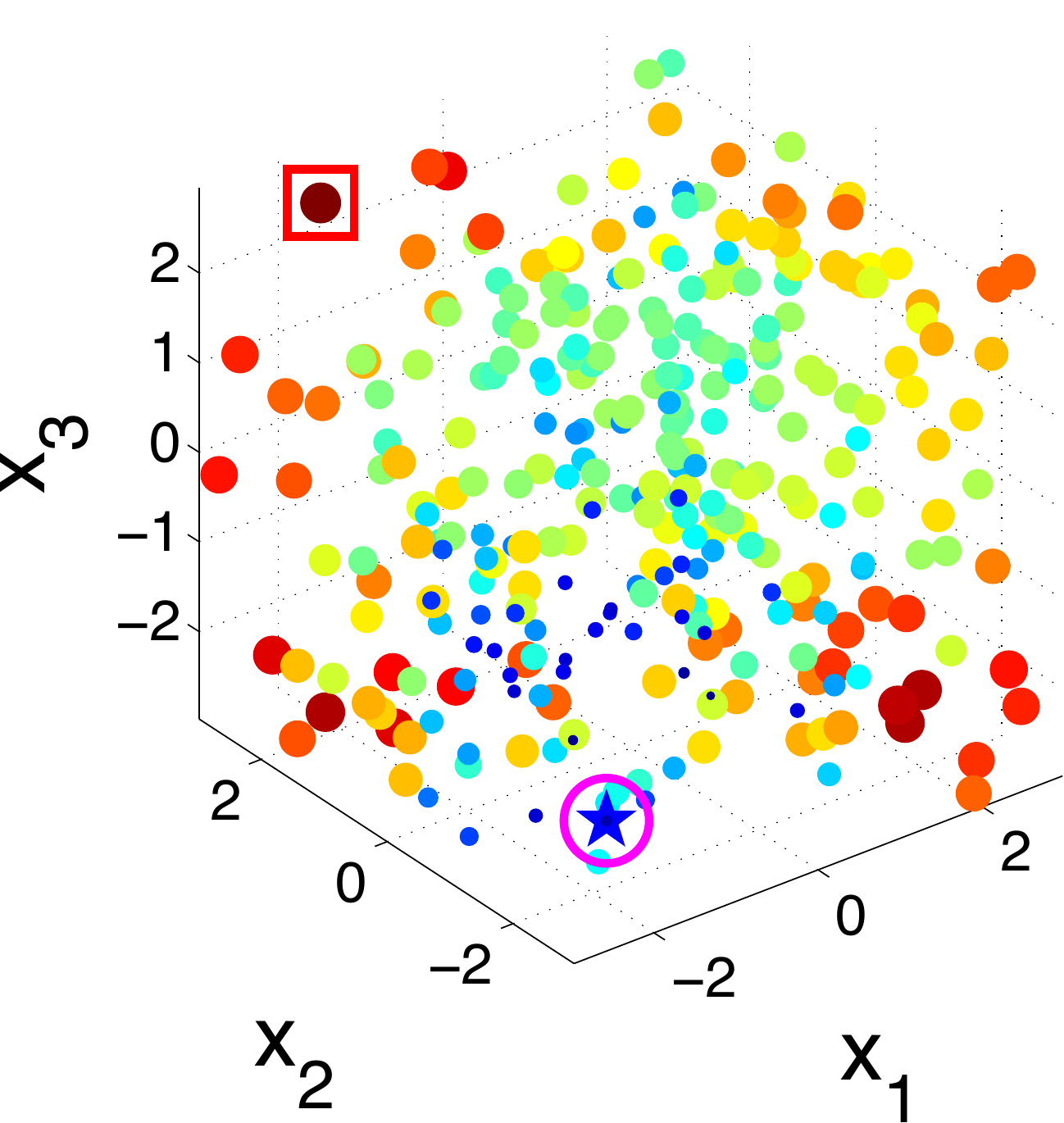} 

    \end{tabular}
    \caption{Snapshots on a $3$-dimensional  dataset. The scatter plots illustrate the KG values at 1-10 iterations from left to right, top to bottom. The star, the red square and pink circle indicate the best alternative, the alternative to be selected and  the implementation decision.  \label{3d}}
\end{figure}

Fig. \ref{3d} illustrates the snapshots of the KG policy on a 3-dimensional synthetic dataset with 300 alternatives. It can be seen that  the KG policy finds the true best alternative after only 10 measurements.  This set of plots also verifies our statement that the KG policy tends to choose data points near the boundary of the region.  
\begin{figure}[hbp!]
\centering
  \subfigure[2-dimensional dataset\label{Abs_a}]{\includegraphics[width=0.335\textwidth]{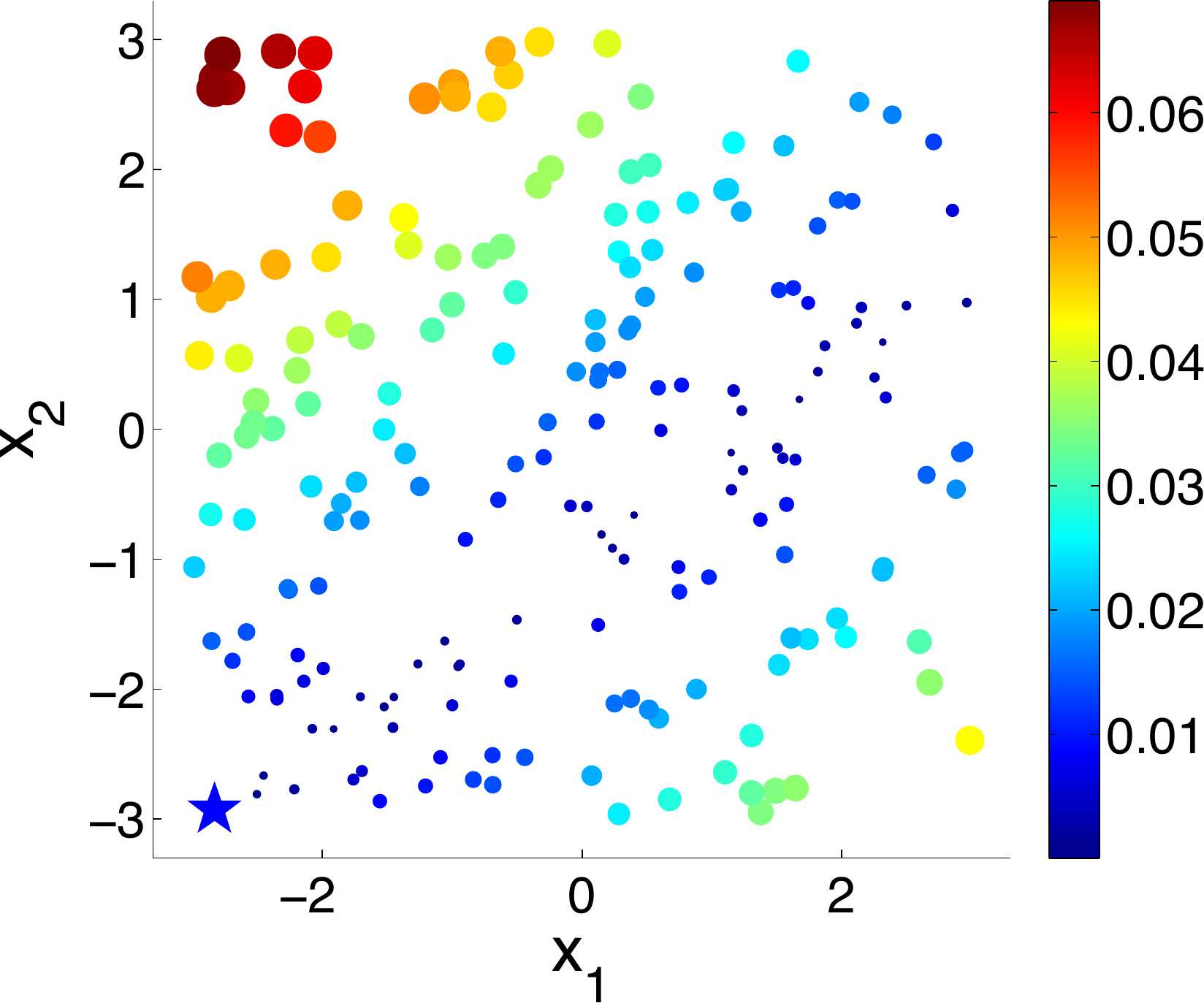}} ~~~~~   
  \subfigure[3-dimensional dataset\label{Abs_b}]{   \includegraphics[width=0.335\textwidth]{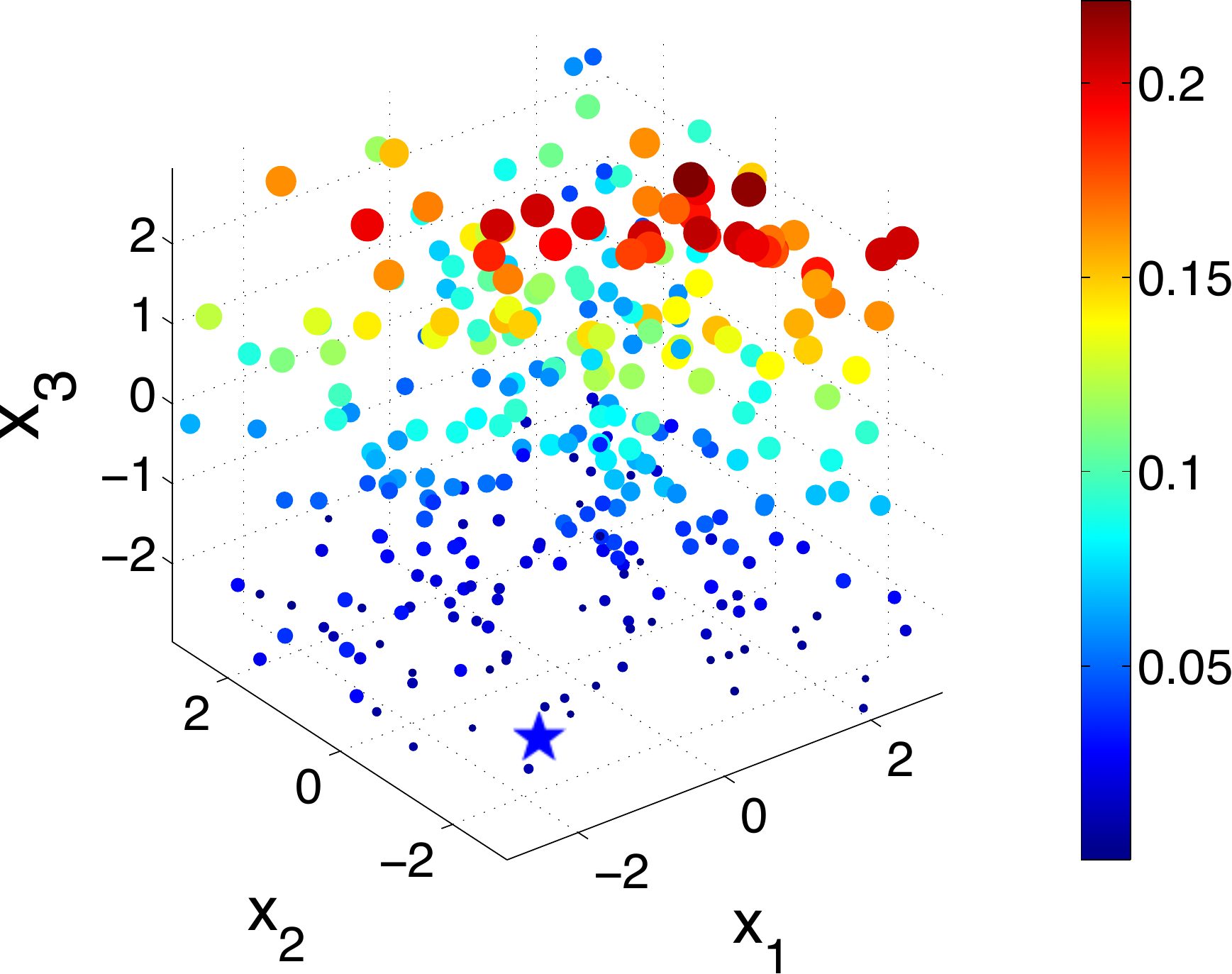}}
\caption{Absolute  error between the predictive probability of $+1$ under current estimate and the true probability. \label{Abs}}
\end{figure}

Also depicted in Fig. \ref{Abs} is the absolute class distribution error of each alternative, which is the absolute difference between the predictive probability of class $+1$ under current estimate and the true probability after $6$ iterations for the 2-dimensional dataset and 10 iterations for the 3-dimensional dataset. We see that the probability  at the true maximum is well approximated, while moderate error in
the estimate is located away from this region of interest.

\subsection{Comparison with other policies} 
Recall that our goal is to maximize the expected response of the implementation decision. We define the Opportunity Cost (OC) metric as the expected response of the implementation decision $\bm{x}^{N+1}:=\arg\max_{\bm{x}}  p(y=+1|\bm{x},\bm{w}^N)$ compared to the true maximal response under weight $\bm{w}^*$: 
$$\text{OC}:=\max_{\bm{x}\in \mathcal{X}} p(y=+1|\bm{x},\bm{w}^*)- p(y=+1|\bm{x}^{N+1},\bm{w}^*).$$
Note that the opportunity cost is always non-negative and the smaller the better.  
To make a fair comparison, on each run, all the time-$N$ labels of all  the alternatives are randomly pre-generated according to the weight vector $\bm{w}^*$ and shared across all the competing policies.   We allow each algorithm to sequentially measure N = 30 alternatives.

\begin{figure*}[t]
   \centering
    \begin{tabular}{ccc}
\subfigure[sonar]{
 \includegraphics[width=0.31\textwidth]{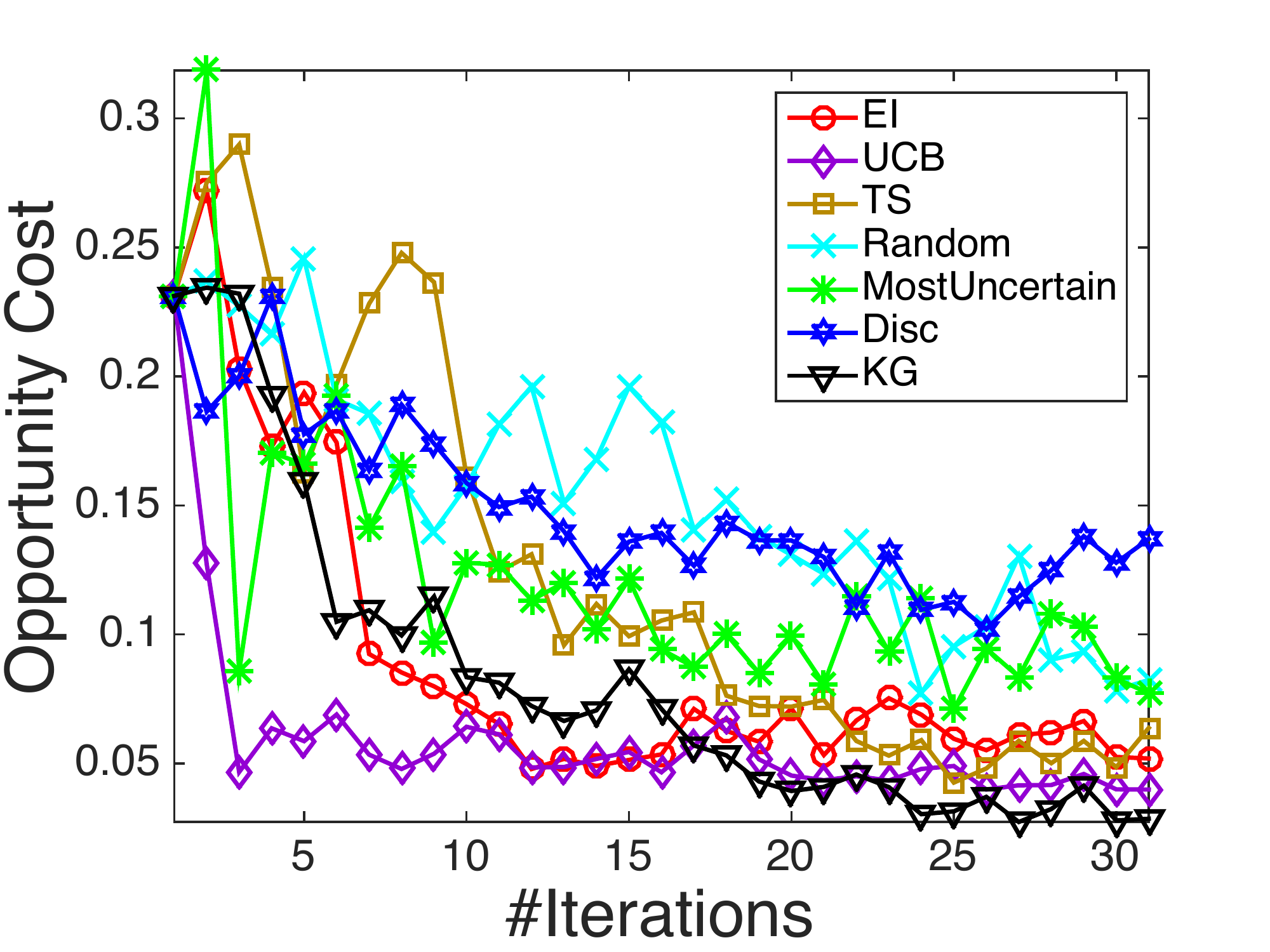} }

 \subfigure[glass identification\label{b}]{
 \includegraphics[width=0.31\textwidth]{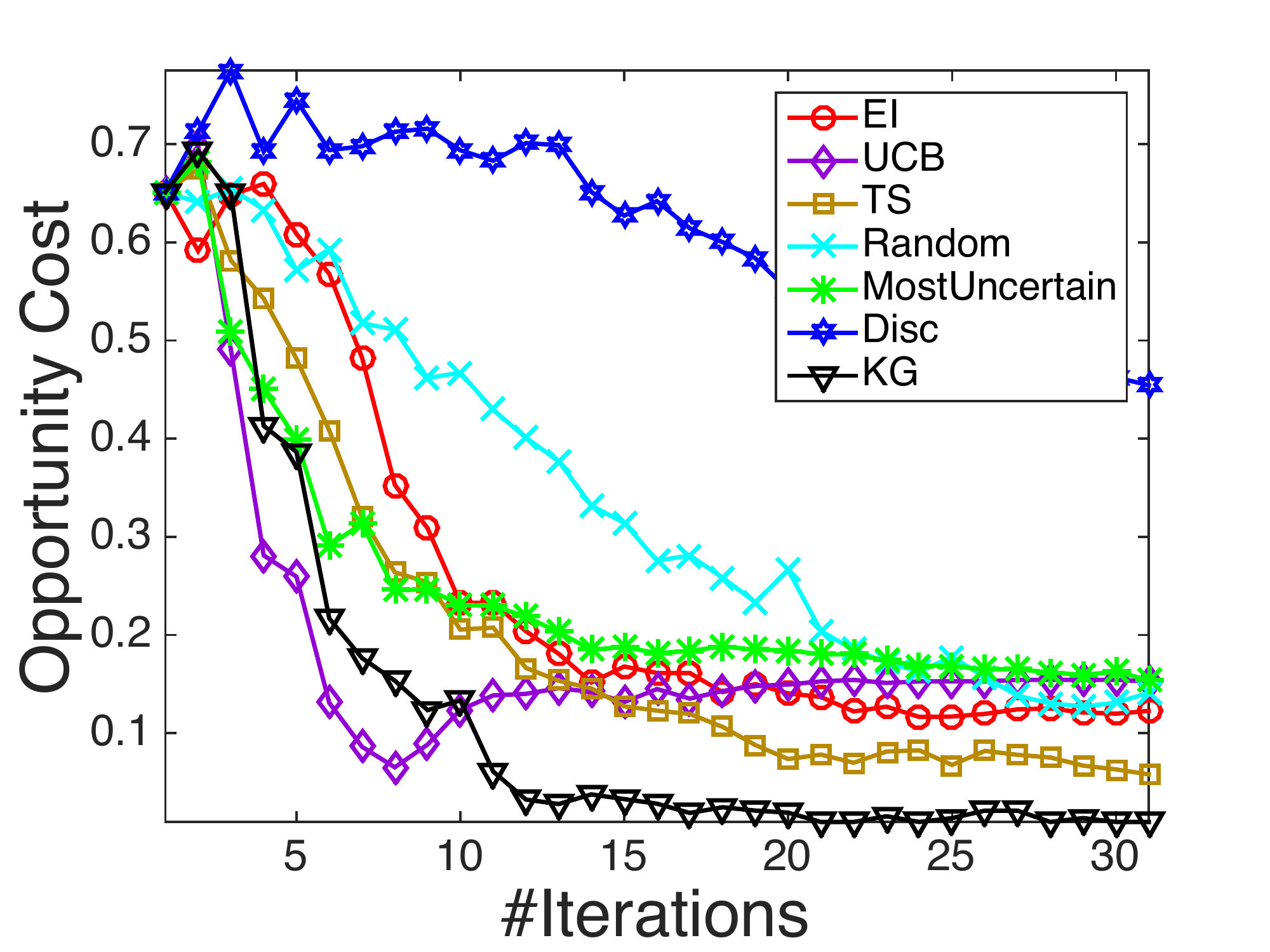} }

  \subfigure[blood transfusion\label{c}]{
 \includegraphics[width=0.31\textwidth]{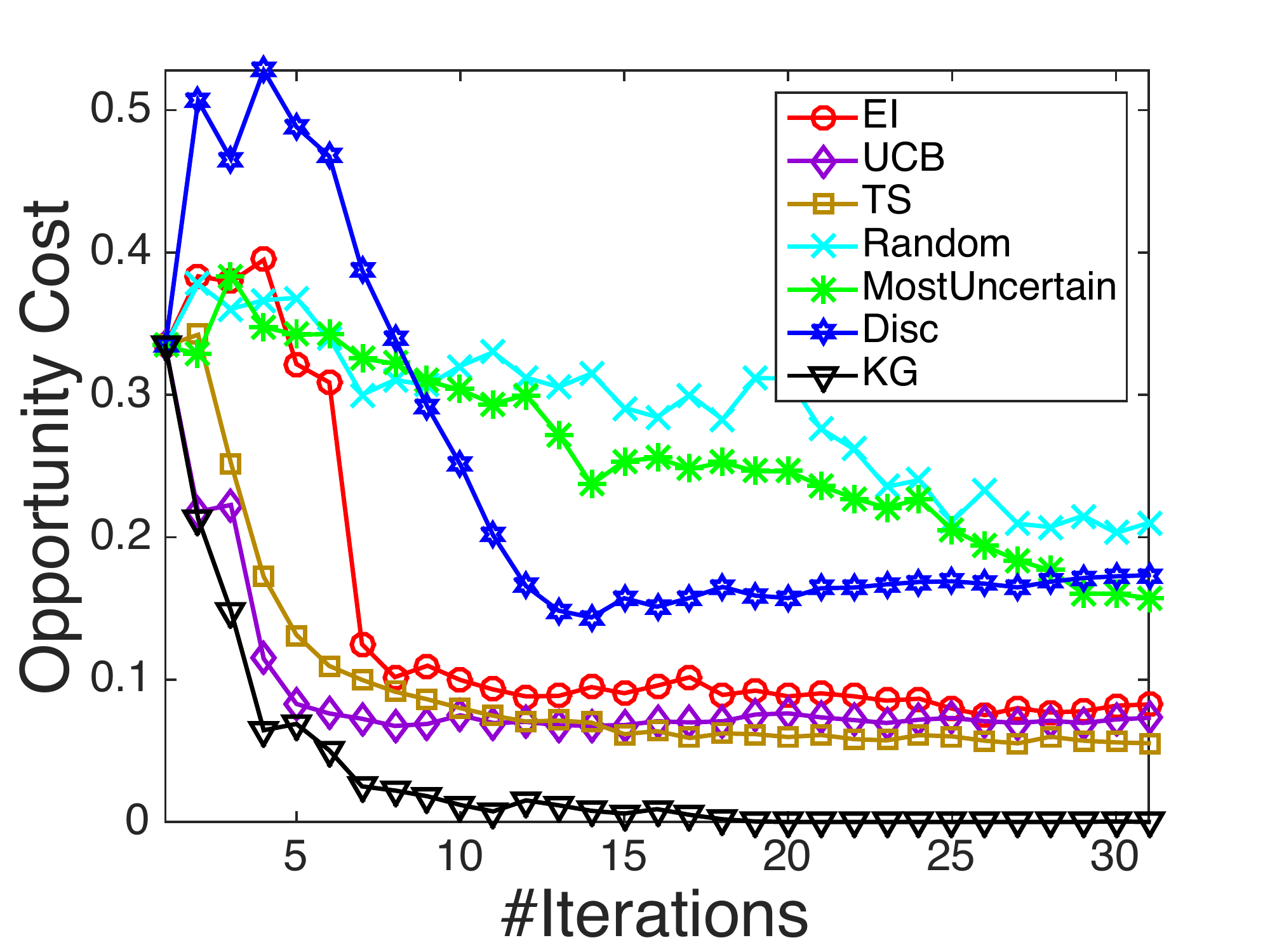} 
 } \\
 \subfigure[survival]{
 \includegraphics[width=0.31\textwidth]{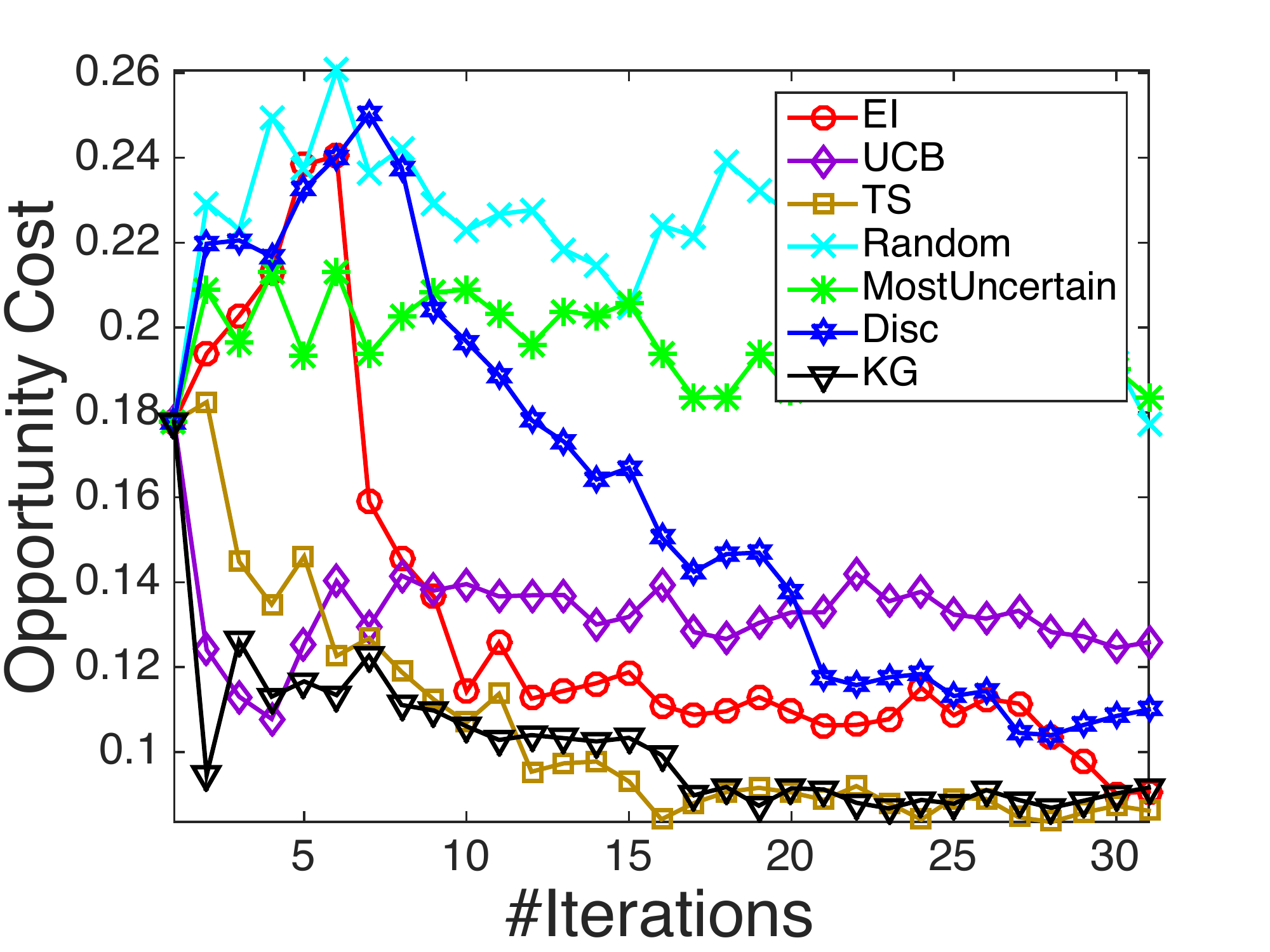} }

 \subfigure[breast cancer (wpbc)\label{e}]{
 \includegraphics[width=0.31\textwidth]{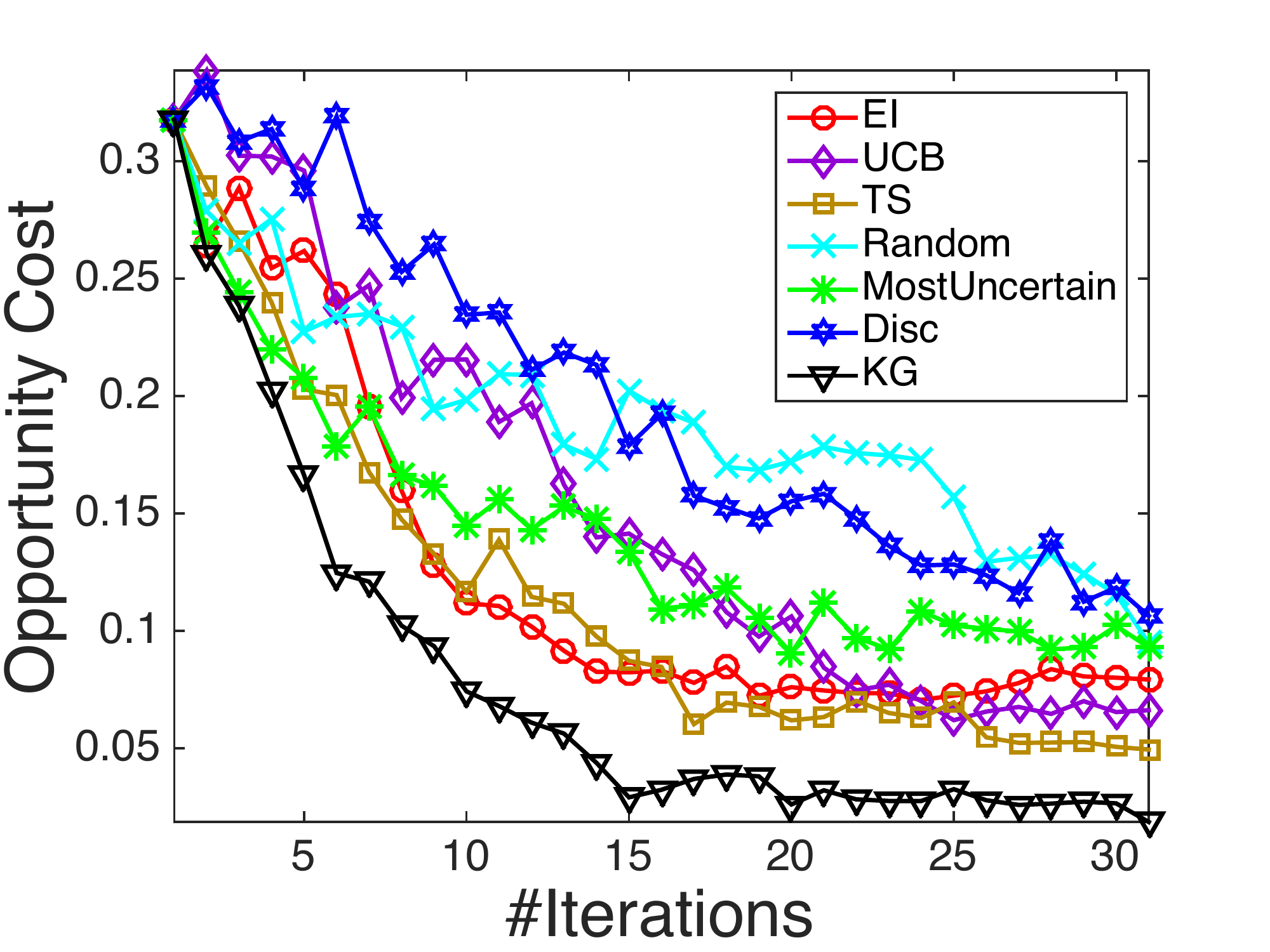} }

  \subfigure[planning relax ]{
 \includegraphics[width=0.31\textwidth]{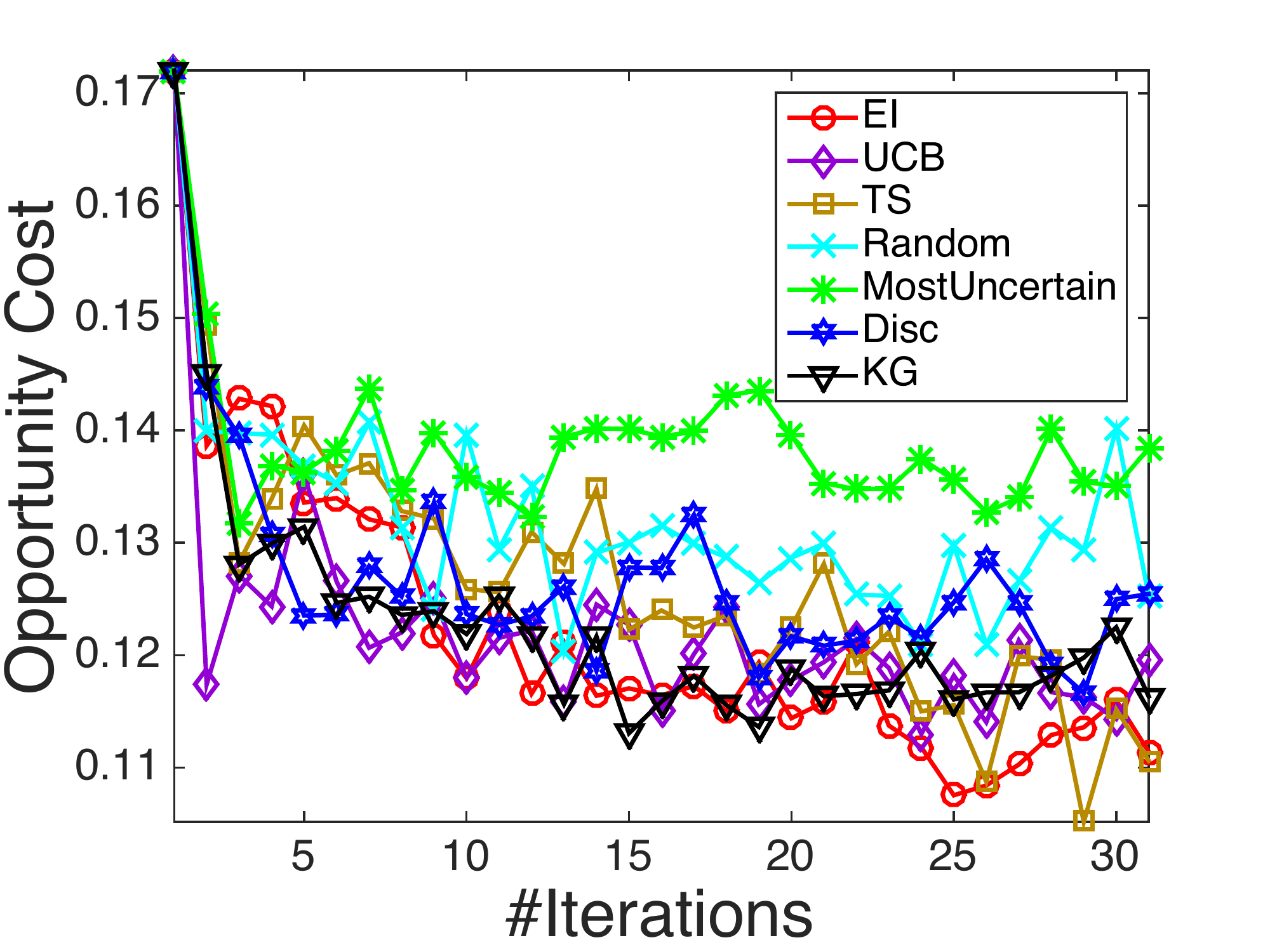} 
 } \\
  \subfigure[climate]{
 \includegraphics[width=0.31\textwidth]{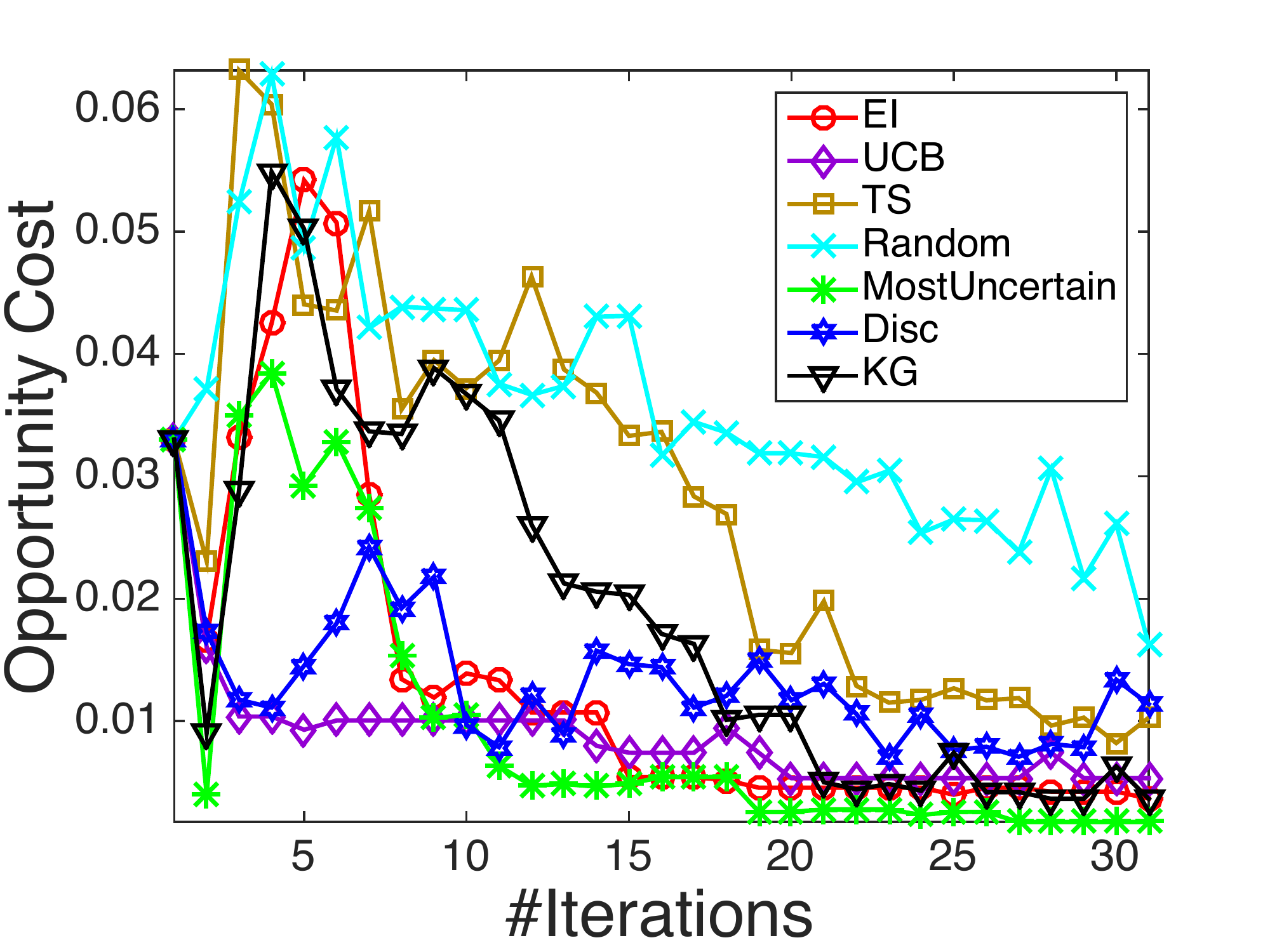} }

 \subfigure[Synthetic data, $d=10$]{
 \includegraphics[width=0.31\textwidth]{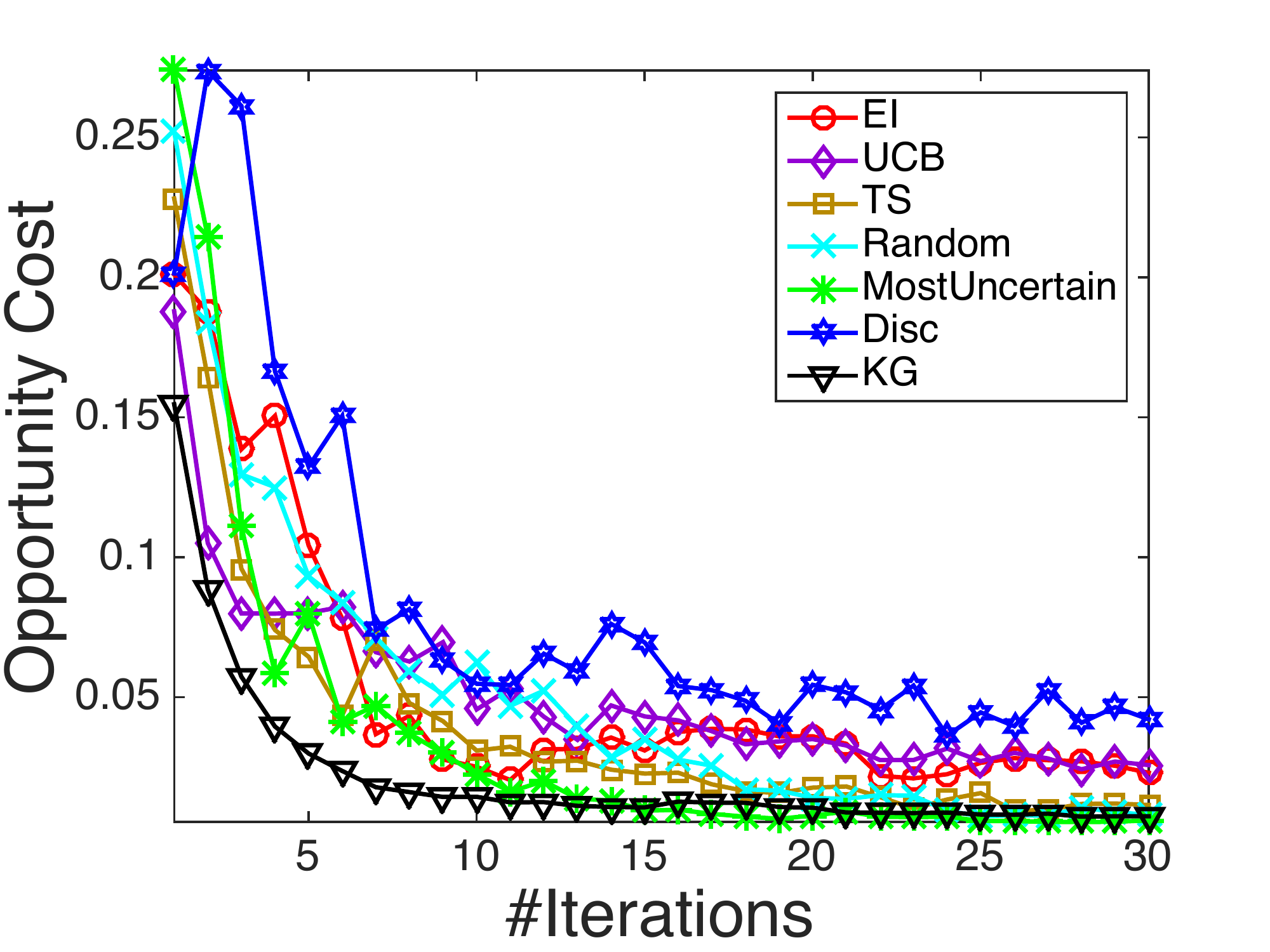} }

  \subfigure[Synthetic data, $d=15$]{
 \includegraphics[width=0.31\textwidth]{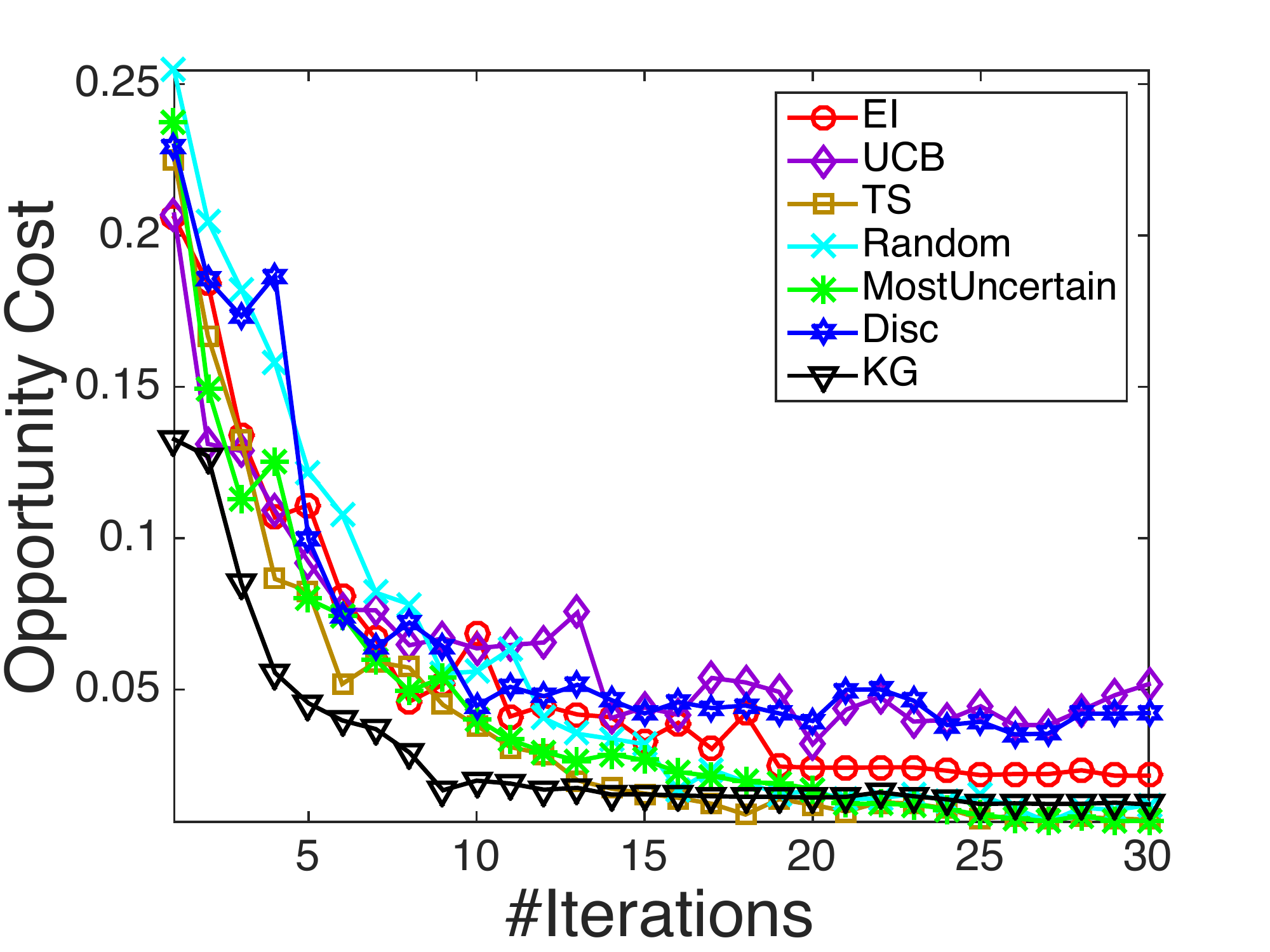} 
 } \\
     \end{tabular}
    \caption{Opportunity cost on UCI and synthetic datasets. \label{33}}
\end{figure*}

 We compare with the following state-of-the-art active learning and Bayesian optimization policies that are compatible with logistic regression: 
Random sampling (Random), a myopic  method that selects the most uncertain instance each step (MostUncertain), discriminative batch-mode active learning (Disc) \cite{guo2008discriminative} with batch size equal to 1, expected improvement (EI) \cite{tesch2013expensive} with an initial fit of 5 examples and Thompson sampling (TS) \cite{chapelle2011empirical}.  Besides, as upper confidence bounds (UCB) methods are often considered in bandit and optimization problems, we compare against UCB on the latent function $\bm{w}^T\bm{x}$ (UCB) \cite{li2010contextual} with $\alpha$ tuned to be 1. All the state transitions are based on the online Bayesian logistic regression framework developed in Section \ref{sec:oblc}, while different policies provides different rules for measurement decisions at each iteration. The experimental results are shown in figure \ref{33}. 
In all the figures, the x-axis denotes the number of measured alternatives and the y-axis represents the averaged opportunity cost  averaged over 100 runs.


It is demonstrated in Fig. \ref{33} that KG outperforms the other policies  in most cases, especially in early iterations, without requiring a tuning parameter. As an unbiased 
selection procedure, random sampling is at least a consistent
algorithm. Yet it is not suitable for expensive experiments where one needs to learn the most  within a small experimental budget.   MostUncertain and Disc perform quite well on  some datasets while badly on others.     A possible explanation is that the goal of active leaning is to  learn a classifier which accurately predicts the labels of new examples so their criteria are not directly related  to maximize the probability of success aside from the intent to learn the prediction.   After enough iterations when active learning methods presumably have the ability to achieve a good estimator of $\bm{w}^*$, their performance will be enhanced. Thompson sampling works in general quite well as reported in other literature \cite{chapelle2011empirical}. Yet, KG has a better performance especially during the early iterations.  In the case when an experiment is expensive and only a small budget is allowed, the KG policy, which is designed specifically to maximize the response, is preferred.  

We also note that KG works better than EI in most cases, especially in Fig. \ref{b}, \ref{c} and \ref{e}.  Although both KG and EI work with the expected value of information,  when EI decides which alternative to measure, it is based on the expected improvement over current predictive posterior distribution while ignoring the potential change of the posterior distribution resulting from the next (stochastic) outcome $y$.  In comparison, KG considers an additional level of expectation over the random (since at the time of decision, we have not yet observed outcome) binary output $y$.

Finally, KG, EI and Thompson sampling outperform the naive use of UCB policies  on the latent function $\bm{w}^T\bm{x}$ due to the errors in the variance introduced by the nonlinear transformation. At each time step,  the posterior of $\log{\frac{p}{1-p}}$ is approximated as a Gaussian distribution.  An upper confidence bound on $\log{\frac{p}{1-p}}$ does not translate to one on $p$ with binary outcomes. In the meantime, KG, EI and Thompson sampling make decisions in the underlying binary outcome probability space and find the right balance of exploration and exploitation.

\section{Conclusion}
Motivated by real world applications, we consider binary classification problems
where we have to sequentially run expensive experiments, forcing us to
learn the most from each experiment. With a small budget
of measurements, the goal is to learn the classification
model as quickly as possible to identify the alternative with
the highest probability of success.  Due to the sequential nature
of this problem, we develop a fast online Bayesian linear
classifier for general response functions to achieve recursive updates. We propose a
knowledge gradient policy using Bayesian linear classification
belief models, for which we use different analytical  approximation
methods to overcome computational challenges. We further extend the  knowledge gradient to the contextual bandit settings. 
We provide a finite-time
analysis on the estimated error, and show that the maximum likelihood estimator based on the adaptively sampled points by the KG policy is consistent and asymptotically normal. We show furthermore that the knowledge gradient policy is asymptotic optimal.  We demonstrate its efficiency through a
series of experiments.

\bibliographystyle{plain}
\bibliography{refer}


\section{Proof of Lemma \ref{eqv}}

First  notice that for any fixed point $\bm{x}$,
\begin{eqnarray*}
\mathbb{E}_N[\text{Pr}(y=+1|\bm{x},\bm{w})]&=&\mathbb{E}_N[\sigma(\bm{w}^T\bm{x})]  \\
&=&\int \sigma(\bm{w}^T\bm{x}) \text{Pr}(\bm{w}|\mathcal{D}^N) \text{d}\bm{w}\\
& =& \text{Pr}(y = +1 | \bm{x},\mathcal{D}^N).
\end{eqnarray*}
By the tower property of conditional expectations, and since $\bm{x}^\pi$ is $\mathcal{F}^N$ measurable,
\begin{eqnarray*}
\mathbb{E}[\text{Pr}(y=+1|\bm{x}^\pi,\bm{w})]&=&\mathbb{E}[\sigma(\bm{w}^T\bm{x}^\pi)] \\
&=&\mathbb{E}\mathbb{E}_N[\sigma(\bm{w}^T\bm{x}^\pi)]\\
&=&\mathbb{E} [\text{Pr}(y = +1 | \bm{x}^\pi,\mathcal{D}^N)].
\end{eqnarray*}
 Then by the definition of $\bm{x}^\pi$, we have $\text{Pr}(y = +1 | \bm{x}^\pi,\mathcal{D}^N)=\max_{\bm{x}} \text{Pr}[y = +1 | \bm{x}, \mathcal{D}^N].$
\hfill 

\section{Proof of Lemma \ref{l2}.} \label{e3}
Let $ f(p)=\sum_{i=1}^d (m_i + yp \frac{x_i}{q_i})x_i$. Then 
\begin{eqnarray*} \frac{\text{d} }{\text{d} { p}} g(p)&=& - \bigg( \frac{\phi(f)^2 }{y\Phi(yf)^2}+\frac{f\phi(f)}{\Phi(yf)}\bigg)\cdot y \sum_{i=1}^d \frac{x_i^2}{q_i} \\
&=& -  \frac{\phi(f)}{\Phi(yf)^2}\sum_{i=1}^d \frac{x_i^2}{q_i}\bigg(\phi(f)+yf\Phi(yf)\bigg).
\end{eqnarray*}

It is obvious that $\frac{\phi(f)}{\Phi(yf)^2}\sum_{i=1}^d \frac{x_i^2}{q_i} \ge 0$. We only need to show that $\phi(f)+yf\Phi(yf) \ge 0$. Since the normal density function $\mathcal{N}(\cdot|0,1)$ is symmetric with respect to the origin and $y \in \{+1, -1\}$, we have $\phi(f)+yf\Phi(yf) = \phi(yf)+yf\Phi(yf)$. Define $u(t)=\phi(t)+t\Phi(t)$. Since   $\lim_{t \to -\infty} u(t)=0$, we show $u(t) \ge 0$ for $t \in [-\infty, \infty]$ by  proving that $u'(t) \ge 0$.

First note that $\frac{\text{d}}{\text{d}t}e^{-t^2/2} = -te^{-t^2/2}$. So we have $\phi'(t)=-t\phi(t)$. Thus $u'(t)=-t\phi(t)+\Phi(t)+t\phi(t) = \Phi(t) \ge 0$.


\section{Proof of Theorem \ref{finite}}

For notation simplicity, we use $\bm{\Sigma}$ and $\bm{m}$ instead of $\bm{\Sigma}^0$ and $\bm{m}^0$,
and use $\bm{x}_i$, $y_i$ to denote $\bm{x}^i$ and $y^i$ in the proof.
Consider the objective function $f(\bm{w})$ defined as 
$$
f(\bm{w}) = \frac{1}{2}(\bm{w}-\bm{m})^T\bm{\Sigma}^{-1}(\bm{w}-\bm{m})  - \mathbb{E}_{\bm{y} \sim \mathcal{B}(\mathcal{D}^n,\bm{w}^*)}\Big{[}\sum_{i=1}^n\log\sigma(y_i\bm{w}^T\bm{x}_i)\Big{]}.
$$
We use $g(\bm{w})$ and $h(\bm{w})$ to denote the quadratic term and the rest, respectively.
By applying the mean value theorem to $h(\bm{w})$, we have
$$h(\bm{w})-h(\bm{w}^*)=(\bm{w} -\bm{w}^*)^T\nabla h(\bm{w}^*)+\frac{1}{2}
(\bm{w} -\bm{w}^*)^TH\big{(}\bm{w}^*+\eta(\bm{w} -\bm{w}^*)\big{)}(\bm{w} -\bm{w}^*),$$
where $H$ is the Hessian of $h(\bm{w})$.
To analyze the first and second order terms in the expansion, we use a similar technique adopted by \cite{ising}.
For the gradient in the first order term, we have
\begin{eqnarray*}
&&\nabla h(\bm{w}^*)=
\mathbb{E}_{\bm{y} \sim \mathcal{B}(\mathcal{D}^n,\bm{w}^*)}
\Big{[}\sum_{i=1}^n \big{(}1-\sigma(y_i\bm{w}^T\bm{x}_i)\big{)}y_i\bm{x}_i\Big{]}\bigg{|}_{\bm{w} = \bm{w}^*}\\
&=&\sum_{i=1}^n \Big{(}\big{(}1-\sigma(\bm{w}^T\bm{x}_i)\big{)}\sigma((\bm{w}^*)^T\bm{x}_i)
-\big{(}1-\sigma(-\bm{w}^T\bm{x}_i)\big{)}\big{(}1-\sigma((\bm{w}^*)^T\bm{x}_i)\big{)}
\Big{)}\bm{x}_i\bigg{|}_{\bm{w} = \bm{w}^*}=0.
\end{eqnarray*}

For the second order term, we have
\begin{eqnarray*}
&&H\big{(}\bm{w}^*+\eta(\bm{w} -\bm{w}^*))\\
&=&\mathbb{E}_{\bm{y} \sim \mathcal{B}(\mathcal{D}^n,\bm{w}^*)}\Bigg{(}\sum_{i=1}^n  \sigma(y_i(\bm{w}^*+\eta(\bm{w} -\bm{w}^*))^T\bm{x}_i)\Big{(}1-\sigma(y_i(\bm{w}^*+\eta(\bm{w} -\bm{w}^*))^T\bm{x}_i)\Big{)}\bm{x}_i\bm{x}_i^T\Bigg{)}\\
&=&\sum_{i=1}^n  \sigma((\bm{w}^*+\eta(\bm{w} -\bm{w}^*))^T\bm{x}_i)\Big{(}1-\sigma((\bm{w}^*+\eta(\bm{w} -\bm{w}^*))^T\bm{x}_i)\Big{)}\bm{x}_i\bm{x}_i^T\\
&=&\sum_{i=1}^n   J_i(\bm{w},\eta)\bm{x}_i\bm{x}_i^T,
\end{eqnarray*}
where $J_i(\bm{w},\eta)=\sigma((\bm{w}^*+\eta(\bm{w} -\bm{w}^*))^T\bm{x}_i)\Big{(}1-\sigma((\bm{w}^*+\eta(\bm{w} -\bm{w}^*))^T\bm{x}_i)\Big{)}$.

We expand $J_i(\bm{w},\eta)$ to its first order and use mean value theorem again,
\begin{eqnarray*}
|J_i(\bm{w},\eta)-J_i(\bm{w}^*,\eta)|&=&|\eta(\bm{w}-\bm{w}^*)^T\bm{x}_i J_i' |\\
&\le&|\sigma(1-\sigma)(1-2\sigma)|  |(\bm{w}-\bm{w}^*)^T\bm{x}_i|\\
&\le&\|\bm{w}-\bm{w}^*\|_2,
\end{eqnarray*}
where we omit the dependence of $\sigma$ on $(\bm{w}^*+\eta(\bm{w} -\bm{w}^*))^T\bm{x}_i$ for simplicity and use the fact $\sigma\in(0,1)$.

Combining the first order and second order analysis and denoting $\|\bm{w} -\bm{w}^*\|_2$ as $R$, we have
\begin{eqnarray*}
h(\bm{w} )-h(\bm{w}^*)&\ge& 
\frac{1}{2}\|\bm{w} -\bm{w}^*\|_2^2  \lambda_{min}
\Big{(}H\big{(}\bm{w}^*+\eta(\bm{w} -\bm{w}^*)\big{)}\Big{)}\\
&\ge& \frac{1}{2}R^2  \lambda_{min}
\Big{(}\sum_{i=1}^n \left [ \sigma(\bm{x}_i^T\bm{w}^*)\big{(}1-\sigma(\bm{x}_i^T\bm{w}^*)\big{)}-R\right ]\bm{x_}i\bm{x}_i^T \Big{)}\\
&\ge& \frac{1}{2}R^2  \lambda_{min}
\Big{(}\sum_{i=1}^n \sigma(\bm{x}_i^T\bm{w}^*)\big{(}1-\sigma(\bm{x}_i^T\bm{w}^*)\big{)}\bm{x_}i\bm{x}_i^T \Big{)} - \frac{1}{2}R^3
\end{eqnarray*}
where we use the fact that $\bm{x}_i\bm{x}_i^T$ is positive semi-definite and $\|\bm{x}_i\|_2\le 1$.
By using $C_{min}$ to denote the minimal eigenvalue in the last inequality, 
we have 
$$h(\bm{w} )-h(\bm{w}^*) \ge \frac{1}{2}C_{min}R^2-\frac{1}{2}R^3.$$

On the other hand, for the quadratic term $g(\bm{w})$, we have
\begin{eqnarray*}
g(\bm{w})-g(\bm{w}^*)&=&(\bm{w}-\bm{w}^*)^T\bm{\Sigma}^{-1}(\bm{w}^*-\bm{m})+\frac{1}{2}(\bm{w}-\bm{w}^*)^T\bm{\Sigma}^{-1}(\bm{w}-\bm{w}^*)\\
&\ge& -DR\sqrt{\lambda_{max}\big{(}\bm{\Sigma}^{-1}\big{)}}+\frac{1}{2}\lambda_{min}\big{(}\bm{\Sigma}^{-1}\big{)}R^2,
\end{eqnarray*}
where$D=\|\bm{\Sigma}^{-\frac{1}{2}}(\bm{w}^*-\bm{m})\|_2$. 

Now define function $F(\Delta)$ as $F(\Delta)=f(\bm{w}^*+\Delta)-f(\bm{w}^*)$,
then we have
\begin{eqnarray}\label{eq1}
F(\Delta)&=&g(\bm{w})-g(\bm{w}^*)+h(\bm{w}^n)-h(\bm{w}^*)\nonumber\\
&\ge& -\sqrt{\lambda_{max}\big{(}\bm{\Sigma}^{-1}\big{)}}DR+\frac{1}{2}\lambda_{min}\big{(}\bm{\Sigma}^{-1}\big{)}R^2+\frac{1}{2}C_{min}R^2-\frac{1}{2}R^3.
\end{eqnarray}
From now on we will use the simplified symbols $\lambda_{max}$ and $\lambda_{min}$ instead of 
$\lambda_{min}\big{(}\bm{\Sigma}^{-1}\big{)} $ and $\lambda_{max}\big{(}\bm{\Sigma}^{-1}\big{)}$.
We consider the sign of $F(\Delta)$ in the case when  $\|\Delta\|_2=\frac{1}{2}(C_{min}+\lambda_{min})$ and
\begin{equation}\label{bound}
D\le\frac{1}{16}\frac{\lambda_{min}^2}{\sqrt{\lambda_{max}}}.
\end{equation}
Based on the analysis in \eqref{eq1}, we have
\begin{eqnarray*}
F(\Delta)&\ge& -\frac{1}{16}\lambda_{min}^2R+\frac{1}{2}\lambda_{min}R^2+\frac{1}{2}C_{min}R^2-\frac{1}{2}R^3\\
&=& \frac{R}{2}\left(-R^2 + (\lambda_{min} + C_{min}) R - \frac{1}{8}\lambda_{min}^2 \right)\\
&=& \frac{R}{2}\left(R^2  - \frac{1}{8}\lambda_{min}^2 \right) > \frac{\lambda_{min}^3}{32}.
\end{eqnarray*}

Notice that $F(0)=0$ and recall that $\bm{w}^n$ minimizes $f(\bm{w})$ so we have 
$$F(\bm{w}^n-\bm{w}^*)=f(\bm{w}^n)-f(\bm{w}^*)\le 0.$$
Then based on the convexity of $F$ we know that $\|\bm{w}^n-\bm{w}^*\|_2\le \frac{1}{2}(C_{min}+\lambda_{min})$, otherwise the values of $F$ at 0, $\bm{w}^n-\bm{w}^*$ and the intersection between the all $\|\Delta\|_2=\frac{1}{2}(C_{min}+\lambda_{min})$ and line from 0 to  $\bm{w}^n-\bm{w}^*$ form a concave pattern, which is contradictory.
Similarly, based on the assumption that $f(\hat{\bm{w}}^n)\le \min_{\bm{w}} f(\bm{w}) + \lambda_{min}^3/32$,
we have 
$$F(\hat{\bm{w}}^n-\bm{w}^*)=f(\hat{\bm{w}}^n)-f(\bm{w}^*)\le \frac{\lambda_{min}^3}{32}.$$
Recall that $F(0)=0$ and $F(\Delta) > \lambda_{min}^3/32$, then the convexity of $F$ also ensures that $\|\hat{\bm{w}}^n-\bm{w}^*\|_2\le \frac{1}{2}(C_{min}+\lambda_{min})$.

Now we start to calculate the probability for \eqref{bound} to hold.
Recall that $\bm{w}^*$ has a prior distribution $\bm{w}^*\sim \mathcal{N}(\bm{m},\bm{\Sigma})$.
Then by denoting the right hand side of  Eq. (\ref{bound}) as $M$, we have
\begin{eqnarray*}
&&\text{Pr}\Big{(}D\le\frac{1}{16}\frac{\lambda_{min}^2}{\sqrt{\lambda_{max}}}\Big{)}\\&=&
\int_{\|\bm{\Sigma}^{-\frac{1}{2}}(\bm{a}-\bm{m})\|\le M}(2\pi)^{-\frac{d}{2}}|\bm{\Sigma}|^{-\frac{1}{2}}\exp \big{(}-\frac{1}{2}(\bm{a}-\bm{m})^T\bm{\Sigma}^{-1}(\bm{a}-\bm{m})\big{)}\mathrm{d}\bm{a}\\
&=&
\int_{\|\bm{b}\|_2\le M}(2\pi)^{-\frac{d}{2}}\exp \big{(}-\frac{1}{2}\bm{b}^T\bm{b}\big{)}\mathrm{d}\bm{b},\\
\end{eqnarray*}
which is the probability of a d-dimension standard normal random variable appears in the ball with radius $M$, $P_d(M)$. This completes the proof.
\hfill 

\section{Proofs of Asymptotic Optimality}\label{e4}
In this section, we provide detailed proofs of all the asymptotic optimal results in Section \ref{5.3}.

\subsection{Proof of Proposition \ref{p1}}
We use $q(\bm{w})$ to denote the predictive distribution under the state $s$ and use $p(\bm{w}|s,\bm{x},y_{\bm{x}})$ to denote the posterior distribution after we observe the outcome of $\bm{x}$ to be $y$.
By Jensen's inequality, we have
\begin{eqnarray*}
\mu_{\bm{x}}^{\text{KG}}(s) &=& \mathbb{E}\Big{[}\max_{\bm{x}'}p(y=+1| \bm{x}',T(s,\bm{x},y_{\bm{x}}))\Big{]}  - \max_{\bm{x}'} p(y=+1|\bm{x}',s) \\
&\ge& \max_{\bm{x}'}\mathbb{E}\big[p(y=+1| \bm{x}',T(s,\bm{x},y_{\bm{x}}))\big]  - \max_{\bm{x}'} p(y=+1|\bm{x}',s).
\end{eqnarray*}
We then show that $\mathbb{E}\big[p(y=+1| \bm{x}',T(s,\bm{x},y_{\bm{x}}))\big] =p(y=+1|\bm{x}',s)$ for any $\bm{x}, \bm{x}'$ and $s$, which leads to $\mu_{\bm{x}}^{\text{KG}}(s) \ge 0$.  Since $y_{\bm{x}}$ is binomial distributed with mean $p(y_{\bm{x}}=+1|\bm{x},s)$, we have
\begin{eqnarray*}
 \hspace*{-5mm} &&\mathbb{E}\big[p(y=+1| \bm{x}',T(s,\bm{x},y_{\bm{x}}))\big] \\
 \hspace*{-4mm} & = &p(y_{\bm{x}}=+1|\bm{x},s)p(y=+1| \bm{x}',T(s,\bm{x},+1)) \\&+& (1-p(y_{\bm{x}}=+1|\bm{x},s))p(y=+1| \bm{x}',T(s,\bm{x},-1)). 
\end{eqnarray*}

Recall that  $p(y=+1|\bm{x},s) 
= \int \sigma(\bm{w}^T\bm{x})p(\bm{w}|s)\text{d}\bm{w}.$  By  Bayes' Theorem, the posterior  distribution in the updated state $T(s,\bm{x},y_x)$ becomes
$$p(\bm{w}'|T(s,\bm{x},+1))=\frac{\sigma((\bm{w}')^T\bm{x})p(\bm{w}'|s)}{\int\sigma(\bm{w}^T\bm{x})p(\bm{w}|s)\text{d}\bm{w}},$$
and
$$p(\bm{w}'|T(s,\bm{x},-1))=\frac{\left (1-\sigma((\bm{w}')^T\bm{x})\right)p(\bm{w}'|s)}{\int\left (1-\sigma(\bm{w}^T\bm{x})\right)p(\bm{w}|s)\text{d}\bm{w}}.$$
Notice that 
\begin{eqnarray*}
p(y=+1|\bm{x}',T(s,\bm{x},+1))&=&\int \sigma((\bm{w}')^T\bm{x}')p(\bm{w}'|T(s,\bm{x},+1))\text{d}\bm{w}'\\
&=&\int \sigma((\bm{w}')^T\bm{x}')
\frac{\sigma((\bm{w}')^T\bm{x})p(\bm{w}'|s)}{\int\sigma(\bm{w}^T\bm{x})p(\bm{w}|s)\text{d}\bm{w}}
\text{d}\bm{w}'\\
&=&\frac{\int \sigma((\bm{w}')^T\bm{x}')\sigma((\bm{w}')^T\bm{x})p(\bm{w}'|s)\text{d}\bm{w}'}
{\int\sigma(\bm{w}^T\bm{x})p(\bm{w}|s)\text{d}\bm{w}},
\end{eqnarray*}
and similarly, we have
$$
p(y=+1|\bm{x}',T(s,\bm{x},-1))=
\frac{\int \sigma((\bm{w}')^T\bm{x}')(1-\sigma((\bm{w}')^T\bm{x}))p(\bm{w}'|s)\text{d}\bm{w}'}
{\int(1-\sigma(\bm{w}^T\bm{x}))p(\bm{w}|s)\text{d}\bm{w}}.
$$
Therefore, 
\begin{eqnarray*}
&&\mathbb{E}\big[p(y=+1| \bm{x}',T(s,\bm{x},y_{\bm{x}}))\big]\\
&=&p(y=+1|s,\bm{x})p(y=+1|\bm{x}',T(s,\bm{x},+1))+
p(y=-1|s,\bm{x})p(y=+1|\bm{x}',T(s,\bm{x},-1))\\
&=&\int \sigma((\bm{w}')^T\bm{x}')\sigma((\bm{w}')^T\bm{x})p(\bm{w}'|s)\text{d}\bm{w}'+
\int \sigma((\bm{w}')^T\bm{x}')(1-\sigma((\bm{w}')^T\bm{x}))p(\bm{w}'|s)\text{d}\bm{w}'\\
&=&\int \sigma((\bm{w}')^T\bm{x}')p(\bm{w}'|s)\text{d}\bm{w}'\\
&=&p(y=+1|s,\bm{x}'),
\end{eqnarray*}
and thus we obtain 
$$\mathbb{E}\big[p(y=+1| \bm{x}',T(s,\bm{x},y_{\bm{x}}))\big]=p(y=+1|s,\bm{x}').$$

\subsection{Proof of Proposition \ref{infty}} \label{eaaa}
The proof is similar to that by \cite{frazier2009knowledge} with additional tricks for Bernoulli distributed random variables. Let $\mathcal{G}$ be the sigma-algebra by the collection $\{\hat{y}^{n+1} \bm{1}_{\{\bm{x}^n = \bm{x}\}}\}$. Since if the  policy $\pi$ measures alternative $\bm{x}$ infinitely often, this collection is an infinite sequence of independent random variables with common Bernoulli distribution with mean $\sigma(\bm{w}^T\bm{x})$, the strong law of large numbers implies $\sigma(\bm{w}^T\bm{x}) \in \mathcal{G}$. Since $\mathcal{G} \in \mathcal{F}^\infty$, we have $\sigma(\bm{w}^T\bm{x}) \in \mathcal{F}^\infty$. Let $U$ be a uniform random variable in $[0,1]$. Then the Bernoulli random variable $y_{\bm{x}}$ can be rewritten as $\bm{1}_{U \le \sigma(\bm{w}^T\bm{x})}$. Since $U$ is independent with $\mathcal{F}^\infty$ and the $\sigma$-algebra generated by $\sigma(\bm{w}^T\bm{x})$, it can be shown that by properties of conditional expectations, 
$$\mathbb{E}\big{[} \sigma(\bm{w}^T\bm{x}')|\mathcal{F}^\infty, \bm{1}_{U \le \sigma(\bm{w}^T\bm{x})}\big{]} = \mathbb{E}\big{[} \sigma(\bm{w}^T\bm{x}')|\mathcal{F}^\infty\big{]}.$$
We next show that the knowledge gradient value of measuring alternative $\bm{x}$ is zero by substituting this relation into the definition of the knowledge gradient. We have
\begin{eqnarray*}
\nu_{\bm{x}}(\mathcal{F}^\infty) &=& \mathbb{E}\bigg{[}\max_{\bm{x}'}\mathbb{E}\big{[} \sigma(\bm{w}^T\bm{x}')|\mathcal{F}^\infty, \bm{1}_{U \le \sigma(\bm{w}^T\bm{x})}\big{]} |\mathcal{F}^\infty \bigg{]} - \max_{\bm{x}'}\mathbb{E}\big{[} \sigma(\bm{w}^T\bm{x}')|\mathcal{F}^\infty \big{]} \\
& =& \mathbb{E} \bigg{[}\max_{\bm{x}'}\mathbb{E}\big{[} \sigma(\bm{w}^T\bm{x}')|\mathcal{F}^\infty \big{]}|\mathcal{F}^\infty \bigg{]} -\max_{\bm{x}'}\mathbb{E}\big{[} \sigma(\bm{w}^T\bm{x}')|\mathcal{F}^\infty \big{]}  \\
&=& \max_{\bm{x}'}\mathbb{E}\big{[} \sigma(\bm{w}^T\bm{x}')|\mathcal{F}^\infty \big{]}  - \max_{\bm{x}'}\mathbb{E}\big{[} \sigma(\bm{w}^T\bm{x}')|\mathcal{F}^\infty \big{]} \\
&=& 0.
\end{eqnarray*}

\subsection{Proof of the Theorem \ref{consistency}: Consistency of the KG Policy}\label{e5}

It has been established that almost surely the knowledge gradient policy will achieve a state (at time T) that the KG values for all the alternatives are smaller than $\epsilon$, after which the probability of selecting each alternative is $1/M$ where $M$ is the number of alternatives.
Notice that in each round, KG policy first picks one out of $M$ alternatives, then a feedback of either 1 or -1 is observed. Equivalently, we can interpret the two procedures  as one of the $2M$ possible outcomes from the set $\tilde{\mathcal{Y}}:=\{(0,\dots, +1/-1,\dots 0) \}$, where each $\tilde{y} \in\tilde{\mathcal{Y}}$ is a $M$-dimensional vector with only element being +1 or -1. It should be noted that by changing the feedback schema in this way will not affect the Bayesian update equations because the likelihood function and the normalization factor in the posterior will both multiply by a factor of $1/M$.  On the other hand, this combined feedback schema makes it possible to treat each measurement $(\bm{x}^n,y^n)$ as  i.i.d. samples in $\tilde{\mathcal{Y}}$.

Define the K-L neiborhood  as  $K_\epsilon(u)=\{v:KL(u,v)<\epsilon\}$, where the K-L divergence is defined as
$KL(u,v):=\int v\log(v/u)$. Since the prior distribution is Gaussian with positive definite covariance matrix, and the likelihood function is the sigmoid function which only takes positive values, then after time $T$, the  posterior probability in the K-L neighborhood of $w^*$ is positive. Based on standard results on the consistency of Bayes' estimates \cite{ghosal2006posterior, ghosal1999posterior,tokdar2007posterior, freedman1963asymptotic}, the posterior is weakly consistent at $w^*$ in the sense that for any neighborhood $U$ of $w^*$, the probability that $\mu(w)$ lies in $U$ converges to 1.
$$\mathbb{P}[U|\tilde{y}^1,\tilde{y}^2,\dots,\tilde{y}^n]\rightarrow 1.$$

Without loss of generality, assume that the alternative $\bm{x}^*$ with the largest probability of +1 is unique, which means $\sigma((\bm{x}^*)^T\bm{w}^*)>\sigma(\bm{x}^T\bm{w}^*)$ for any alternative $\bm{x}$ other than $\bm{x}^*$. Then we can pick $U$ to be the neighborhood of $\bm{w}^*$ such that $\sigma((\bm{x}^*)^T\bm{w})>\sigma(\bm{x}^T\bm{w})$ holds for any $\bm{w}\in U$. 
The neighborhood $U$ exists because we only have finite number of altervatives.
From the consistency results, the probability that the best arm under  posterior estimation is the true best alternative goes to 1 as the measurement budget goes to infinity.

\end{document}